%% file: main.tex
\documentclass{article}
\usepackage{iclr2027_conference,times}

\usepackage[T1]{fontenc}
\usepackage{amsmath,amssymb}
\usepackage{graphicx}
\usepackage{booktabs}
\usepackage{xcolor}
\usepackage[skins]{tcolorbox}
\usepackage{float}
\usepackage{algorithm}
\usepackage{algpseudocode}
\usepackage{caption}
\usepackage{setspace}
\definecolor{citeblue}{RGB}{32,126,190}
\usepackage[breaklinks,colorlinks,citecolor=citeblue]{hyperref}
\usepackage{url}
\usepackage{xspace}

\definecolor{ppaccent}{HTML}{2E6E8E}
\iclrfinalcopy

\newcommand{\system}{LiteReality-Agent\xspace}

\definecolor{gallerylabel}{HTML}{F2D58A}
\definecolor{objectpanel}{HTML}{E9EEF3}

\title{LiteReality-Agent: An Agentic System for Interactable 3D Indoor Scene Reconstruction}
\newcommand{\paperauthorblock}{%
\parbox[t]{\dimexpr\textwidth-2\tabcolsep\relax}{%
  \centering\normalfont
  {\bfseries
    Zhening~Huang$^{1*\dagger}$\quad Yueyan~Li$^{2*}$\quad
    Johnathan~Chiu$^{3*}$\quad Xiaoyang~Lyu$^1$\quad Matt~Zhou$^3$\par
    \vspace{2pt}
    Yuxin~Yao$^1$\quad Joan~Lasenby$^1$\quad Shangzhe~Wu$^1$\par}
  \vspace{4pt}
  {\small $^1$University of Cambridge\quad
    $^2$Imperial College London\quad $^3$Independent Researcher\par}
  \vspace{3pt}
  {\small
    \href{https://litereality.github.io/agent/}{\texttt{litereality.github.io/agent/}}\par}
}
}
\author{\paperauthorblock}
\newcommand{\papercontributionnote}{%
  \begingroup
  \renewcommand{\thefootnote}{}%
  \begin{NoHyper}
  \footnotetext{$^*$Equal Contribution.\quad $^{\dagger}$Project Lead.}%
  \end{NoHyper}
  \endgroup
}
\hypersetup{pdfauthor={Zhening Huang, Yueyan Li, Johnathan Chiu, Xiaoyang Lyu, Matt Zhou, Yuxin Yao, Joan Lasenby, Shangzhe Wu}}

\graphicspath{{figures/}}

\begin{document}
\maketitle
\fancyhead{}
\renewcommand{\headrulewidth}{0pt}
\papercontributionnote

\begin{center}
  \refstepcounter{figure}
  \includegraphics[width=0.88\linewidth]{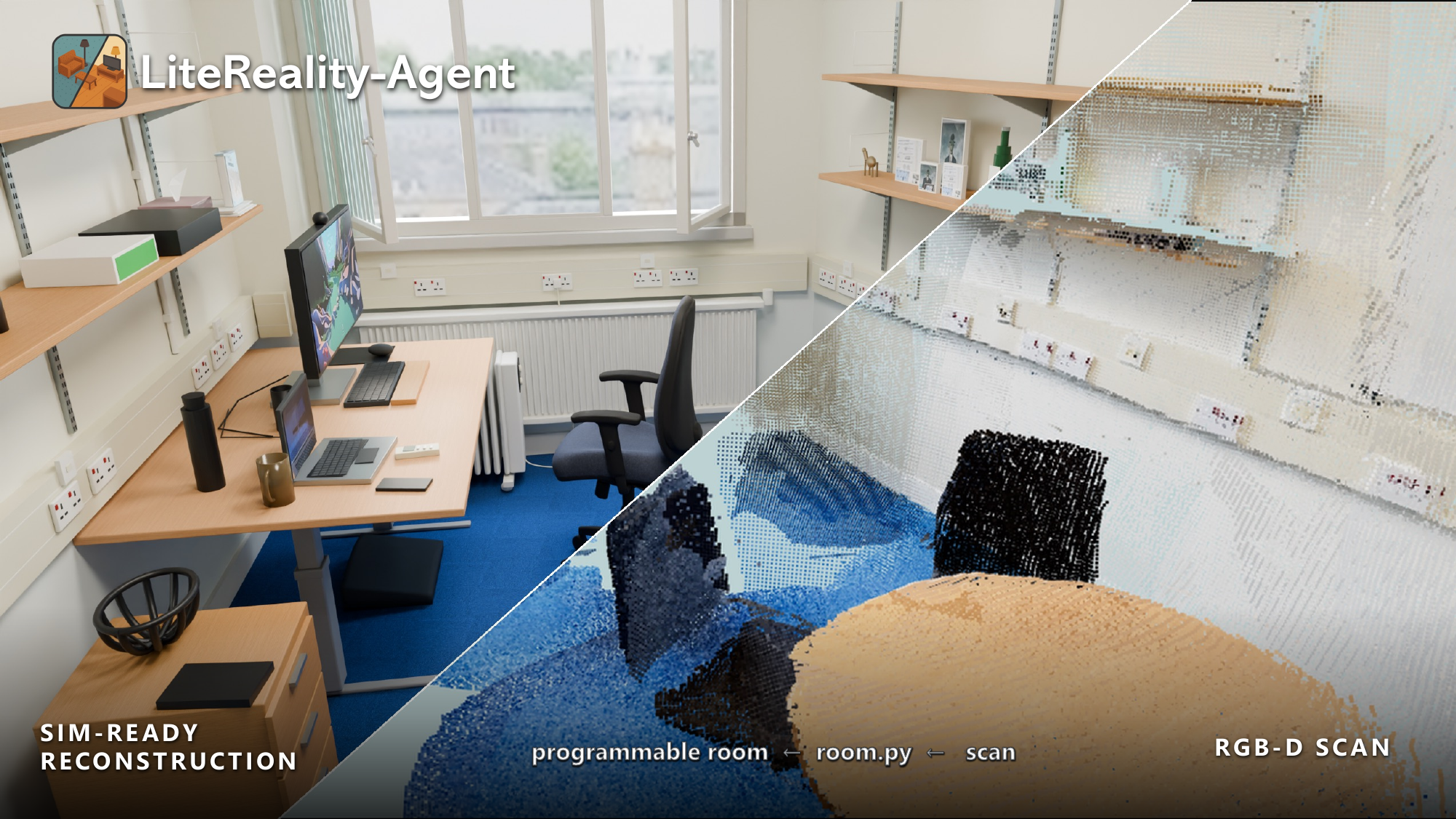}
  \par\vspace{0.35em}
\parbox{0.95\linewidth}{\small\textbf{Figure~\thefigure: Reconstructing a real room with \system{}.} \system{} repurposes the reconstruction as a coding task, where a coding agent writes a \texttt{Room.py} file that can be compiled into a realistic, interactive, and simulation-ready 3D room.}
    
  \label{fig:teaser}
\end{center}

\input{sections/0_abstract}
\input{sections/1_introduction}
\input{sections/2_related_work}
\input{sections/3_litereality_agent}

\input{sections/4_results}

\input{sections/5_conclusion}

\clearpage

\input{sections/6_acknowledgments}
\bibliographystyle{iclr2027_conference}
\bibliography{references}

\clearpage
\appendix
\setcounter{figure}{0}
\renewcommand{\thefigure}{A\arabic{figure}}
\input{sections/appendix/appendix}

\end{document}

%% file: sections/0_abstract.tex
\begin{abstract}
We present \system{}, an agentic system for reconstructing real indoor environments as realistic, articulated, and simulation-ready 3D scenes from RGB-D scans. At its core, \system{} formulates 3D reconstruction as a coding problem, where a coding agent gathers evidence through designed tools
and iteratively edits a Python script, \texttt{Room.py}, which can be executed to produce a 3D digital twin of the room.
With this formulation, we build a robust observe-edit-verify harness that supports evidence gathering, measurement, verification, layout optimisation, simulation-readiness, and quality control throughout the reconstruction process. \system{} produces high-quality reconstructions suitable for simulation and downstream embodied AI tasks. Furthermore, as agent capabilities continue to improve rapidly, the system introduced by \system{} remains a strong orchestration framework for agents: it equips them with specialised tools, structured workflows, and robust verification mechanisms that substantially improve reconstruction quality and reliability. We have demonstrated that \system{} produces reconstructions that are more geometrically accurate, visually realistic, and simulation-compatible than those generated by the latest frontier models, such as Astra and Fable. We therefore view \system{} as a practical and important building block for robust real-to-sim systems. Both the source code and the data-capture application are publicly available.
\end{abstract}

%% file: sections/1_introduction.tex
\section{Introduction}
\label{sec}
Building a functional, simulation-ready digital twin has long been a holy grail of inverse graphics. Achieving this goal could unlock a range of applications, including robotic simulation, manufacturing digital twins, and virtual reality. Unlike point-based 3D reconstruction methods \citep{campos2021orbslam3, wang2025vggt} , or conventional mesh
reconstruction methods \citep{murez2020atlas,sun2021neuralrecon}, our goal is to recover a structured environment that can be directly used for interaction and physical simulation \citep{huang2026litereality}. This requires jointly understanding the structure of the scene, reconstructing objects and their articulations \citep{zhou2026articraft,li2026instructparticulate}, estimating material and physical properties, and resolving spatial relationships. Integrating these capabilities into a robust end-to-end system remains a longstanding challenge.

Previous work has taken this direction through several approaches. CAD-based scene reconstruction retrieves CAD models and positions them within the scene to form a compact representation \citep{avetisyan2018scan2cad,kuo2020mask2cad,gao2024diffcad}. Compositional 3D reconstruction from single or multiple images instead uses 3D generative models to reconstruct individual objects and place them in the scene \citep{yao2025cast}, while inverse-rendering pipelines recover object geometry, orientation, and materials through differentiable rendering \citep{yan2023psdrroom,yeh2022photoscene}. Despite this progress, a gap remains in building a robust and scalable system that can turn large-scale real-world scans into high-quality environments suitable for interaction and physical simulation. 
LiteReality \citep{huang2026litereality} takes a step toward this goal with a multi-stage pipeline that converts indoor scans into graphics-ready scenes by integrating scene understanding and parsing, asset retrieval, appearance recovery, and assembly into a conventional graphics representation. However, its largely deterministic pipeline limits its ability to handle cases beyond predefined scenarios.

Meanwhile, a growing body of work has explored the use of large language models for 3D scene generation. Systems such as Holodeck \citep{yang2024holodeck}, SceneCraft \citep{hu2024scenecraft}, and SceneSmith \citep{pfaff2026scenesmith} employ LLMs to generate and validate scene layouts and object placements, producing simulation-ready indoor environments in graphics engines. At the object level, Articraft \citep{zhou2026articraft} formulates articulated asset generation as a programming task supported by a domain-specific SDK and automated verification. More recently, proprietary LLMs such as GPT-6 Astra and Claude Fable 5 further suggest that coding agents have developed strong capabilities for understanding and modelling 3D environments. Together, these advances make the previously daunting problem of functional digital-twin reconstruction increasingly tractable. What remains missing is a system that can orchestrate such agents while grounding their outputs in real-world observations and ensuring that the resulting scenes are consistently high quality and suitable for their intended applications.

To this end, we introduce \system{}, a full-stack agentic system for transforming consumer RGB-D room scans into realistic, articulated, and simulation-ready environments. At its core, we formulate reconstruction as a coding task: the entire room is represented as an executable  \texttt{Room.py} program that serves as a persistent workspace. A coding agent can access the observations, modify the program, and compile it into a 3D scene. Because all modifications operate through the same executable representation, the scene remains readable, reproducible, and directly editable throughout the reconstruction process.

\system{} consists of two stages: \textit{Scene Initialisation} and \textit{Agentic Authoring}. \textit{Scene Initialisation} first reconstructs each object from its corresponding image evidence as either a static or articulated asset. The noisy scan-derived layout and object placements are then processed by the \textit{Layout Agent}, which resolves collisions, validates object placements and support relationships, and produces a clean, physically plausible scene layout. The resulting scene is instantiated as an initial \texttt{Room.py} file.
\textit{Agentic Authoring} subsequently refines the scene through an \textbf{observe-edit-verify} harness. The agent is equipped with task-specific tools for image selection, rendering and comparison, geometric and metric measurements, and scene editing. Explicit quality-control checks, including collision and support verification, help ensure that the scene remains complete, structured, and physically plausible throughout the authoring process. 
Physical properties and collision geometry are also estimated by the object reconstruction agents, following a formulation similar to Articraft \citep{zhou2026articraft}. The resulting environment is simulation-ready and fully programmable through \texttt{Room.py}. Representing the entire scene as an executable Python program also enables downstream editing via natural-language instructions carried out by a coding agent. We demonstrate this capability through scene decoration, object replacement, and layout adjustment.

We rigorously benchmark \system{} on diverse real-world indoor scenes against the deterministic LiteReality pipeline and general-purpose coding agents: GPT-6 Astra running in Codex and Fable 5.1 running in Claude Code. Our evaluation combines image-based measures of visual fidelity, geometric measures of reconstruction accuracy and physical plausibility, and human and agent-based perceptual assessments. \system{} achieves the highest mean scores across all reported visual metrics and perceptual criteria. These results demonstrate the effectiveness of \system{} as an orchestration framework for agentic indoor real-to-simulation reconstruction. Both the source code and the data-capture application are publicly available.

%% file: sections/2_related_work.tex
\section{Related Work}
\label{sec:related-work}

\paragraph{Indoor Reconstruction and Real-to-Sim.}
Object-centric methods retrieve and align CAD assets to observations
\citep{avetisyan2018scan2cad,kuo2020mask2cad,gao2024diffcad}, while neural representations
recover scene appearance and geometry~\citep{mildenhallnerf,kerbl20233d,nidecompositional}.
LiteReality~\citep{huang2026litereality} and MetaScenes~\citep{yu2025metascenes}
convert scans into asset-based environments. Video2Game~\citep{xia2024video2game},
DRAWER~\citep{xia2025drawer}, and HoloScene~\citep{xia2025holoscene} reconstruct
interactive worlds from video; DRAWER additionally recovers articulation.
FIRE3D~\citep{xia2026fire3d} predicts compositional scene assets with a feed-forward network.
For robot learning, Phone2Proc~\citep{deitke2022phone2proc} generates procedural
environments from mobile captures, RialTo~\citep{torne2024rialto} constructs digital twins
for policy refinement, and PolaRiS~\citep{jain2025polaris} reconstructs environments
for policy evaluation. ACDC~\citep{dai2024automated} instead constructs digital cousins
using similar assets.

\paragraph{Object Generation and Articulation.}
Image-to-3D methods range from multi-view synthesis
\citep{liu2023zero1to3,liu2024syncdreamer,liuone2345,long2024wonder3d} to large reconstruction
and generative models~\citep{hong2024lrm,xuinstantmesh,xiang2025native,hunyuan3d2025hunyuan3d,sam3dteam2025sam}.
CAST~\citep{yao2025cast} aligns generated objects into a scene, while
URDFormer~\citep{chen2024urdformer} infers articulated simulation environments from images.
Real2Code~\citep{mandi2024real2code} predicts articulation as code;
Articulate-Anything~\citep{le2025articulateanything} iteratively generates and critiques
articulation programs. Articraft~\citep{zhou2026articraft} couples programmatic asset
generation with a specialised SDK and validation harness.

\paragraph{Language-Guided Scene Authoring.}
LayoutGPT~\citep{feng2023layoutgpt} and Holodeck~\citep{yang2024holodeck} generate
layouts from language; FirePlace~\citep{huang2025fireplace} and
LayoutVLM~\citep{sunlayoutvlm} refine placement through geometric or visual feedback.
3D-GPT~\citep{sun2023gpt}, SceneCraft~\citep{hu2024scenecraft}, and
LL3M~\citep{lu2025ll3m} use language models for programmatic 3D authoring.
The Scene Language~\citep{zhang2025scenelanguage} combines programs, words, and embeddings
to represent scene structure and appearance. SceneSmith~\citep{pfaff2026scenesmith},
SAGE~\citep{xia2026sage}, and SceneWeaver~\citep{yang2025sceneweaver} organise agentic
scene-generation pipelines. BlenderAlchemy~\citep{huang2024blenderalchemy} explores
visually guided editing, and BlenderGym~\citep{gu2025blendergym} benchmarks code-based
graphics editing.

\paragraph{Agentic Inverse Graphics.}
SceneScript~\citep{avetisyan2024scenescript} and IG-LLM~\citep{kulits2024igllm}
infer structured scene descriptions from visual inputs. VIGA~\citep{yin2026viga}
iterates through code, rendering, and inspection; Thinking in Blender~\citep{he2026thinking}
refines scene factors in executable code from a single image.
HARMONY~\citep{sun2026harmony} combines agentic placement with geometric refinement
for monocular scene reconstruction. Agentic Real2Sim~\citep{chen2026agenticreal2sim}
uses vision-language agents to assemble physical simulations from interaction videos.
The AHa-3D research blog~\citep{xu2026aha3d} describes a closely related workflow
for video-to-simulation reconstruction with camera, measurement, and inspection tools.
\system{} addresses metric RGB-D room reconstruction, combining articulated assets
and a persistent scene program with calibrated multi-view verification and explicit
collision and support checks.

%% file: sections/3_litereality_agent.tex
\section{\system{}}
\label{sec:litereality-agent}
\system{} reconstructs a room in two stages: \emph{scene initialisation}
establishes the layout and major objects, and \emph{agentic authoring}
refines an executable scene program through an observe--edit--verify
loop (Fig.~\ref{fig:system-overview}). Implementation details and
supporting examples are provided in Appendix~\ref{app:system_details}.

\begin{figure*}[t]
  \centering
  \includegraphics[width=\textwidth]{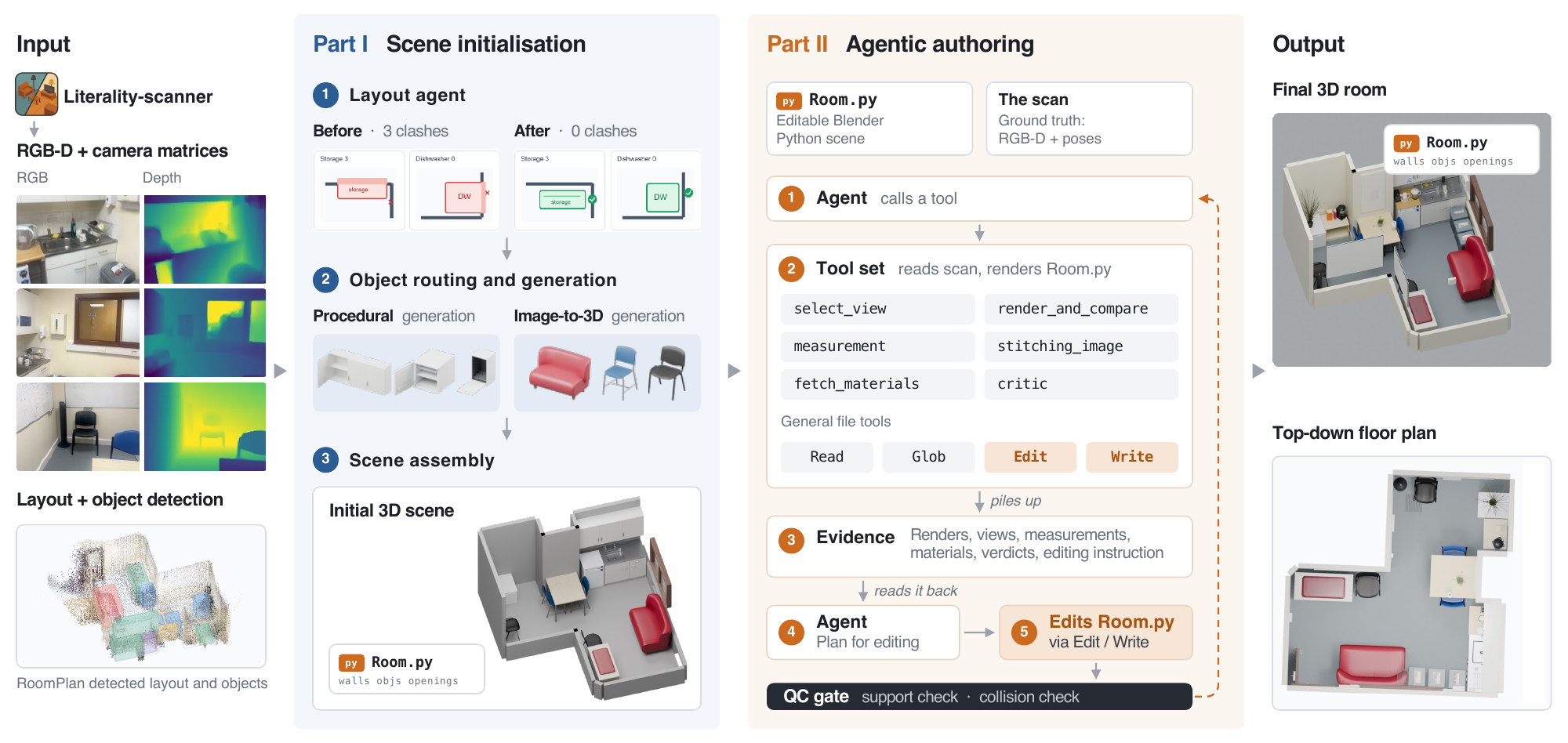}
  \caption{\textbf{End-to-end LiteReality-Agent pipeline.}
  RGB--D scans and detections guide layout repair and object reconstruction
  to produce an executable initial room. The authoring agent refines
  \texttt{Room.py} through an observe--edit--verify loop, using capture
  views, measurements, materials, and rendered comparisons. Control gates
  bound authoring and check geometry and support. The editable scene and
  physical metadata support MuJoCo export.}
  \label{fig:system-overview}
\end{figure*}

\subsection{Input: RGB--D Scans and Detection Results}
\label{sec:problem-setting}
The input consists of RGB--D scans and detection results from the
\emph{LiteReality Scan} application, including the room layout and
object bounding boxes.

\subsection{Scene Initialisation}
\label{sec:initialization}
We follow LiteReality~\citep{huang2026litereality} in parsing the room
shell and reconstructing objects separately before assembly, replacing
asset retrieval with generation from the scan's own images.

\paragraph{Layout Agent.}
The layout is represented as a scene graph of structural elements and
objects, with relations for support, wall adjacency, functional grouping,
and potential clashes. These relations distinguish physical support
from association: a chair belongs to a table but rests on the floor.
They also distinguish incompatible overlaps from expected box overlap,
such as a chair tucked beneath a table.

The Layout Agent corrects noisy detections before object reconstruction.
Geometric checks identify objects outside the room, embedded in walls,
unsupported, or overlapping incompatibly. Deterministic repairs make
small placement and dimension adjustments; image-based reasoning handles
ambiguous cases, such as an incorrectly sized or merged detection.
Proposals are accepted only when they improve geometric consistency and
remain within cumulative limits relative to the scan. The corrected
boxes guide both asset generation and final placement, preventing later
assembly from restoring the original errors.

\paragraph{Object reconstruction.}
For each object, selected capture views are processed by an
image-generation model to produce a clean reference image. Structured or
articulated objects are routed to procedural generation, while visually
complex, predominantly static objects use TRELLIS.2~\citep{xiang2025native}.
Generated meshes are checked for malformed geometry, detached fragments,
and fused background surfaces. A bounded retry loop for generated chairs
regenerates the reference and mesh when these checks fail.

The procedural branch follows Articraft~\citep{zhou2026articraft}: a coding
agent writes an \texttt{object.py} program defining geometry, separate
moving parts, and joints. We compile physical properties alongside the
visual asset, retaining authored values and otherwise estimating mass
from category and size, inertia from geometry, and friction from material
priors. Convex collision geometry and per-object simulation checks help
identify unusable assets before room assembly.

\paragraph{The scene program.}
The shell and reconstructed objects are assembled at their corrected
scales and poses. \texttt{Room.py} defines the room, object instances,
materials, and added geometry, while procedural assets retain their
\texttt{object.py} programs. This is the agent's persistent workspace:
edits are written to code and reproduced by rebuilding the scene.

\subsection{Agentic Authoring}
\label{sec:authoring}
The second stage observes the capture, edits the scene program, and
verifies the result through rendered comparisons and geometric checks.
Algorithm~\ref{alg:authoring} summarises the loop, and
Table~\ref{tab:tools} lists its tools.

\begin{figure*}[t]
\begin{minipage}[t]{0.50\linewidth}
\captionof{algorithm}{Stage 2 agentic authoring loop: the agent iteratively edits \texttt{Room.py} based on observations using the tools in Table~\ref{tab:tools}.}
\label{alg:authoring}
\scriptsize
\begin{tabular}{@{}p{\linewidth}@{}}
\toprule[1pt]
\begin{algorithmic}[1]
\State $\mathit{env}, \mathit{tools} \gets \texttt{Room.py},\; \mathit{defined\_tools}$
\State $\mathit{observations} \gets \Call{Scan}{\,}$
\State $\mathit{history}, \mathit{turn}, \mathit{qc\_passed} \gets [\,],\; 0,\; \textsc{false}$
\While{$\mathit{turn} < T$ \textbf{and not} $\mathit{qc\_passed}$}
    \State $\mathit{actions} \gets \Call{LLM}{\mathit{observations}, \mathit{env}, \mathit{tools}, \mathit{history}}$
    \State $\mathit{result} \gets \Call{Execute}{\mathit{actions}, \mathit{env}, \mathit{observations}}$
    \State append $(\mathit{actions}, \mathit{result})$ to $\mathit{history}$
    \If{$\mathit{result}$ edited the environment}
        \State $\mathit{qc\_passed}, \mathit{feedback} \gets \Call{QC}{\mathit{env}, \mathit{observations}}$
        \State append $\mathit{feedback}$ to $\mathit{history}$
    \EndIf
    \State $\mathit{turn} \gets \mathit{turn} + 1$
\EndWhile
\end{algorithmic}
\\
\bottomrule[1pt]
\end{tabular}
\end{minipage}\hfill
\begin{minipage}[t]{0.47\linewidth}
\captionof{table}{Authoring tools and shared evidence utilities.
View and render tools support room, wall, and object scopes.}
\label{tab:tools}
\scriptsize
\setlength{\tabcolsep}{2pt}
\renewcommand{\arraystretch}{1.05}
\setlength{\aboverulesep}{1pt}
\setlength{\belowrulesep}{1pt}
\begin{tabular}{@{}p{0.34\linewidth} p{0.61\linewidth}@{}}
\toprule[1pt]
Tool / utility & Purpose \\
\midrule[0.5pt]
\texttt{select\_views} & Select informative capture frames for a target \\
\texttt{render\_and\_}\newline\texttt{compare} & Compare calibrated scene renders with photographs \\
\texttt{measurement} & Measure dimensions and positions in metres \\
\texttt{stitch\_wall} & Assemble a rectified wall image and coverage mask \\
\texttt{fetch\_material} & Fetch PBR texture maps and adjust their colour \\
\midrule
\texttt{bash, glob}\newline\texttt{read, edit} & Find files, read code and image evidence, and create or edit programs \\
\bottomrule[1pt]
\end{tabular}
\end{minipage}
\end{figure*}

\subsubsection{Evidence Tools}
\paragraph{Image selection at three levels.}
The agent queries the whole room, one wall, or one object. Room-level
selection greedily chooses complementary frames that cover the walls.
Wall-level selection ranks views by the wall's projected on-screen area.
Object-level selection uses per-object visibility from the render
manifest to find clear views of the target. A minimum frame separation
reduces near-duplicate observations. These scopes let the agent move
from inspecting the overall layout to checking a local detail.

\paragraph{Rendering and comparison.}
The same three levels determine how comparisons are labelled
(Fig.~\ref{fig:authoring-evidence}a--c). Room views label visible objects
by name; wall views outline only the selected wall; object views draw
its bounding box. Each pair uses the same calibrated camera for the
render and photograph. Thus, a room-level mismatch can be traced to a
named object or surface and inspected in isolation. Changed scene and
object programs are rebuilt before rendering so that comparisons show
the current edit.

\paragraph{Wall stitching and measurement.}
Capture frames are projected onto a metric wall plane and combined into
one head-on reference, selecting visible observations while marking
unobserved regions. A grid expresses distance along the wall and height
above the floor. For example, the whiteboard in
Fig.~\ref{fig:app_wall_stitching}d can be located and sized from its edges
relative to the red and blue ticks, then placed using those coordinates
in \texttt{Room.py}. Surface queries also pair this rectified reference
with an orthographic render for checking materials and fixtures. Raw
views remain necessary for objects protruding from the wall. PBR
retrieval supplies texture maps whose colour can be adjusted to match
the observations.

\paragraph{Authoring harness.}
The harness guides four ordered modules: \emph{first}, matching wall
materials; \emph{second}, adding fixtures and lighting; \emph{third},
placing small objects; and \emph{finally}, refining existing procedural
assets through multi-view renders of their geometry, appearance, and
articulation. Each module builds on the previous result. To keep the
scene organised for later editing, added geometry is grouped into named
objects with explicit support (\texttt{rests\_on}) or attachment
(\texttt{attached\_to}) relationships. Subsequent agents can then identify
and edit objects while accounting for these relationships.

\subsubsection{Control and Quality Gates}
\paragraph{Budgeted control.}
Since Visual quality is hard to set reliable automatic stopping criterion, so we
control authoring through a bounded loop with explicit gates. Sessions
have turn and tool-call limits. During the final $10\%$ of calls,
exploration and self-checks are disabled so the agent completes and saves
its edits. The program is then rebuilt and validated. The budget controls
when authoring stops; geometric and support gates determine whether the
result passes validation.

\paragraph{Geometric and support gates.}
\label{sec:quality}
Layout repair uses coarse boxes, so the authored room requires another
check on its actual geometry. This is particularly important for
chair--table pairs with valid box overlap and for newly added objects.
Mesh intersections determine whether a collision exists; convex
approximations estimate corrective translations. Bounded corrections
move unanchored objects, are saved to \texttt{Room.py}, and are checked
again after rebuilding.

Support is checked separately: an object may avoid collisions yet float
above its intended surface. We compare object undersides with declared
supports, report missing supports or invalid contact heights, and check
that articulated assets retain separate moving parts. Unresolved
findings return to authoring; checklist-based visual review can provide
additional appearance feedback. Table~\ref{tab:layout_evaluation} reports
support and intersection measurements on the resulting scenes.

\subsection{Simulation-Ready Export}
\label{sec:simulation-export}
We export the room to MuJoCo~\citep{todorov2012mujoco} as individual objects
with mass, inertia, collision geometry, and joints. Walls and attached
fixtures stay fixed; movable objects have free joints, and articulated
parts use hinge or slide joints. Separate collision shapes preserve space
beneath tables and inside furniture. The exporter accounts for asset
placement and scale, estimating physical properties when needed.

%% file: sections/4_results.tex
\section{Results}
\label{sec:experiments}

We evaluate visual fidelity, geometric consistency, and perceived
similarity on ten indoor scenes, comparing LiteReality-Agent with
LiteReality and two general-purpose agent baselines.

\subsection{Experimental Setup}
\label{sec:experimental_setup}

\paragraph{Scenes, baselines, and inputs.}
We evaluate LiteReality-Agent on ten indoor scenes with 620 capture
views, comparing against LiteReality~\citep{huang2026litereality} and two general-purpose agents:
GPT-6 running in Codex and Fable 5.1 running in Claude Code.
Each general-purpose agent is evaluated with RGB photographs alone
(RGB-only) and with additional geometric observations from
LiteReality Scan (+Scan), including depth, camera calibration,
a point cloud, and a RoomPlan layout. All methods use the same
photographs for each scene; LiteReality and LiteReality-Agent
receive the same scan information as the +Scan baselines.

\paragraph{Evaluation protocol.}
All reconstructions are aligned to the recorded camera coordinate
system without changing scene scale or individual object placements,
and rendered at $960\times720$ using the recorded camera poses and
intrinsics. We evaluate fidelity at the captured viewpoints rather
than held-out novel views. All methods share the lighting and
colour-management settings of the corresponding LiteReality-Agent
scene while retaining their own materials and textures, including
material emission. This lighting is not independently calibrated
to the reference photographs.
Additional input and evaluation details are provided in
Appendix~\ref{app:experimental_details}.

\subsection{Visual Fidelity}
\label{sec:visual_results}

\input{tables/main_results}

\begin{figure*}[t]
    \centering
    \includegraphics[
    width=0.92\linewidth,
    trim=0 280 0 0,
    clip
]{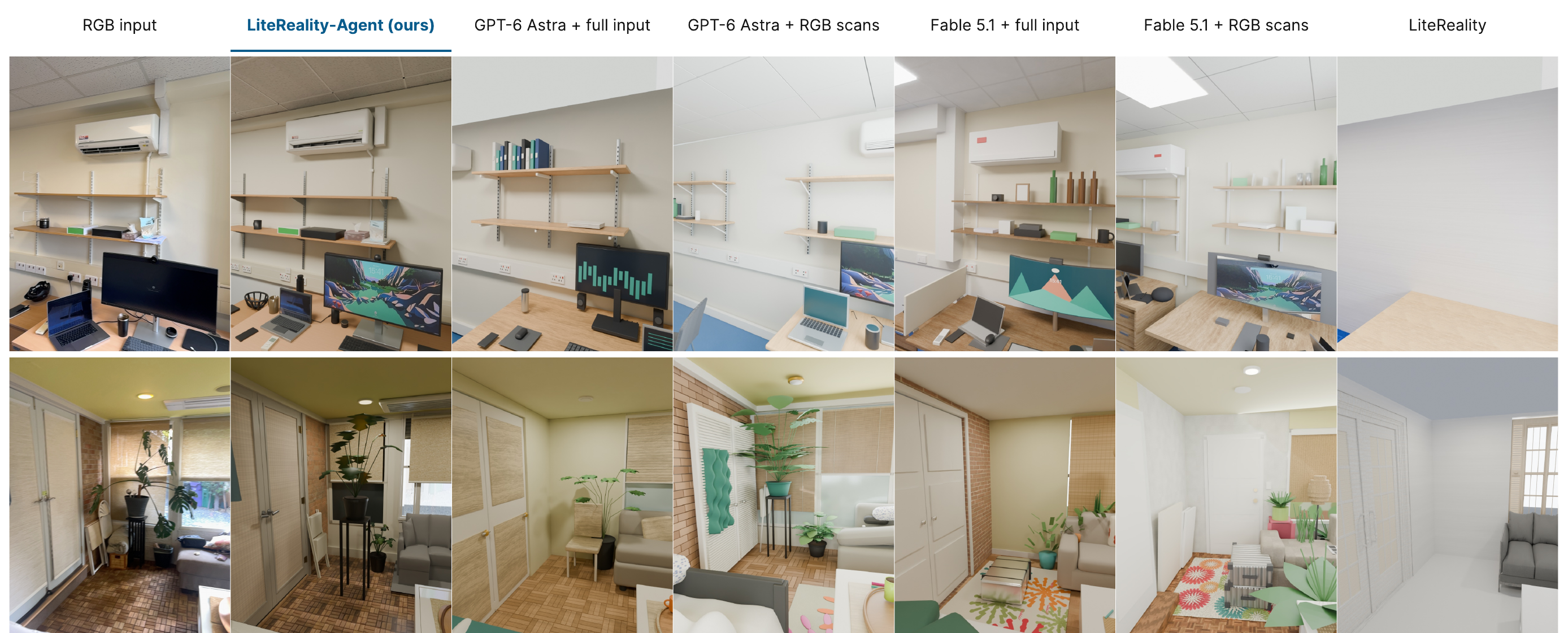}
    \caption{
        \textbf{Qualitative comparison with reconstruction baselines.}
        Three representative scenes compare the RGB input and RGB--D scan
        with \system{}, GPT-6 Astra and Fable 5.1 with and without scan input, and
        LiteReality. All reconstruction results are rendered from the same
        calibrated camera viewpoint.
    }
    \label{fig:baseline-comparison}
\end{figure*}

We evaluate SSIM, PSNR, RMSE, and LPIPS on reference photographs
and renders paired at the same resolution, without additional
image alignment or appearance correction. We retain all 620
capture views and average per-frame scores with equal weight,
so scenes with more views contribute more to the aggregate;
metric configurations are provided in
Appendix~\ref{app:experimental_details}.
As shown in Table~\ref{tab:visual_results}, LiteReality-Agent
achieves the best mean on all four metrics. Compared with
GPT-6 (+Scan), the strongest alternative, it improves PSNR by
$1.162$\,dB and reduces LPIPS by $13.7\%$; compared with
LiteReality, the corresponding gains are $2.248$\,dB and
$19.6\%$. Figure~\ref{fig:baseline-comparison} presents qualitative
comparisons on three representative scenes.

\subsection{Geometric Consistency and Physical Plausibility}
\label{sec:physics_results}

We evaluate three aspects of geometric consistency:
depth MAE (cm) over 611 frames with shared valid depth,
bottom near-contact (the proportion of furniture objects with
at least one sampled bottom point within $1$\,cm of an external
supporting surface), and the number of furniture pairs with
intersecting triangle surfaces.
Furniture counts vary across methods, so object-based results
must be interpreted alongside these counts.
These measures assess specific aspects of physical plausibility,
rather than establishing full physical validity.
Implementation details are provided in
Appendix~\ref{app:geometric_details}.

\paragraph{Results.}
Table~\ref{tab:layout_evaluation} shows that LiteReality-Agent
outperforms both scan-informed agent baselines on all three
criteria, achieving a depth MAE of $10.50$\,cm
($21.2\%$ lower than GPT-6 (+Scan)), bottom near-contact for all
86 identified furniture objects, and no detected
furniture--furniture surface intersections.
Compared with LiteReality, it maintains $100\%$ bottom
near-contact while reducing depth MAE from $15.03$ to
$10.50$\,cm and detected intersection pairs from 43 to zero.

\subsection{Effect of Scan Data on General-Purpose Agents}
\label{sec:scan_input}

Adding LiteReality Scan data to the same RGB photographs improves
all four visual metrics for both general-purpose agents.
PSNR increases from $11.549$ to $12.967$\,dB for GPT-6 and
from $10.681$ to $12.287$\,dB for Fable 5.1.
Depth MAE decreases from $40.51$ to $13.33$\,cm and
from $58.64$ to $16.32$\,cm, respectively, corresponding to
reductions of $67.1\%$ and $72.2\%$.

These fidelity gains do not consistently extend to the
object-based layout measures: detected furniture intersection
pairs increase from 14 to 24 for GPT-6 and from 36 to 54 for
Fable 5.1, while GPT-6's bottom near-contact rate decreases
from $69.1\%$ to $63.0\%$.
These comparisons support the complete system but do not isolate the
harness from its other components; intersection counts also depend on
the differing furniture populations.

\subsection{Qualitative Results}
\label{sec:case-studies}

\begin{figure}[t]
    \centering
    \includegraphics[width=0.90\linewidth]{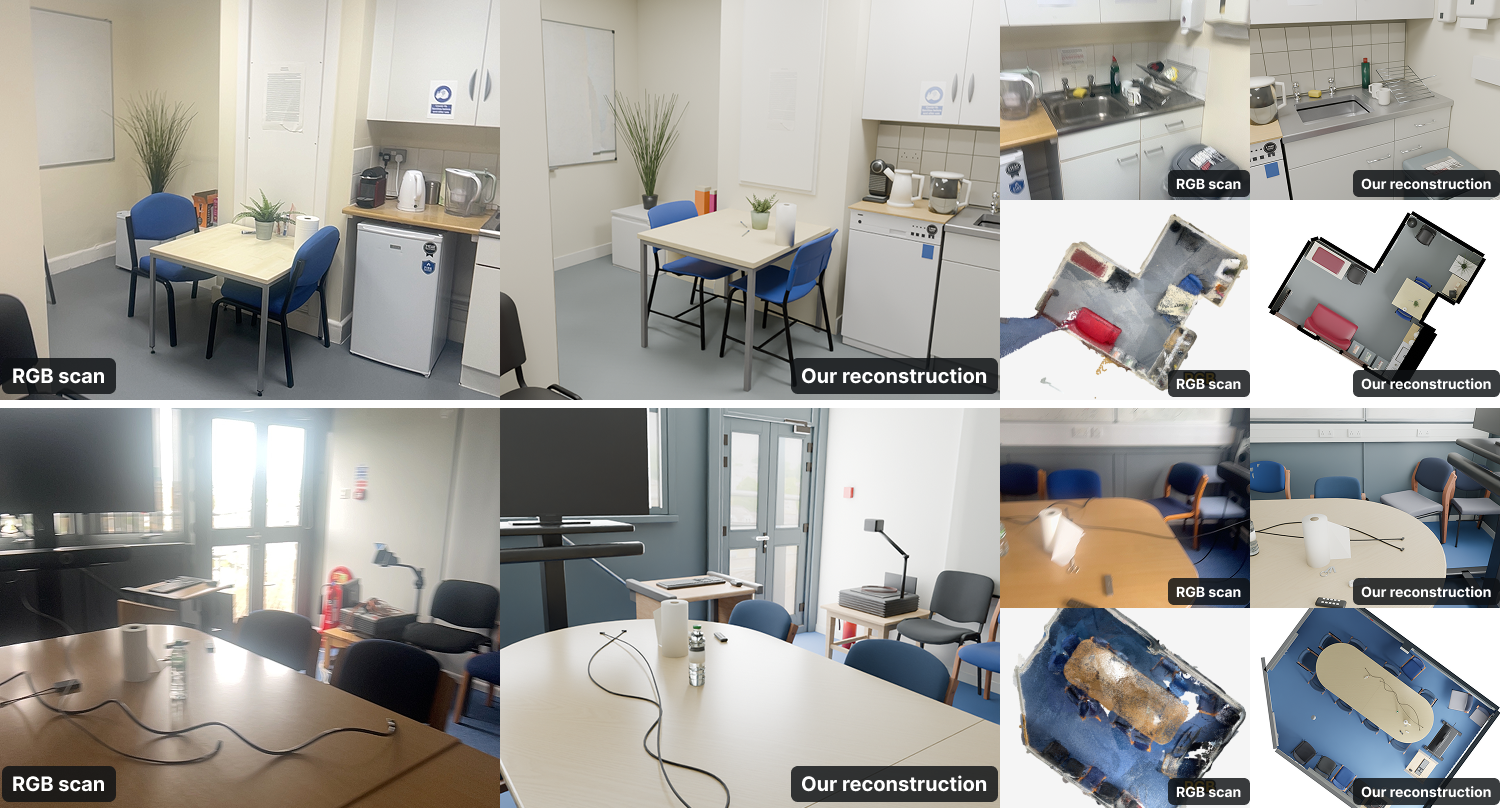}
    \caption{
        \textbf{Representative reconstruction comparisons.}
        A kitchen (top) and meeting room (bottom), with captured RGB
        photographs, reconstructed views, and point-cloud/scene plan views.
    }
    \label{fig:cases}
\end{figure}

\begin{figure}[t]
\centering
\begin{minipage}{0.98\linewidth}
\begin{minipage}[t]{0.485\linewidth}
  \vspace{0pt}
  \centering
  \begin{minipage}[t]{.49\linewidth}
    \centering
    \includegraphics[width=\linewidth]{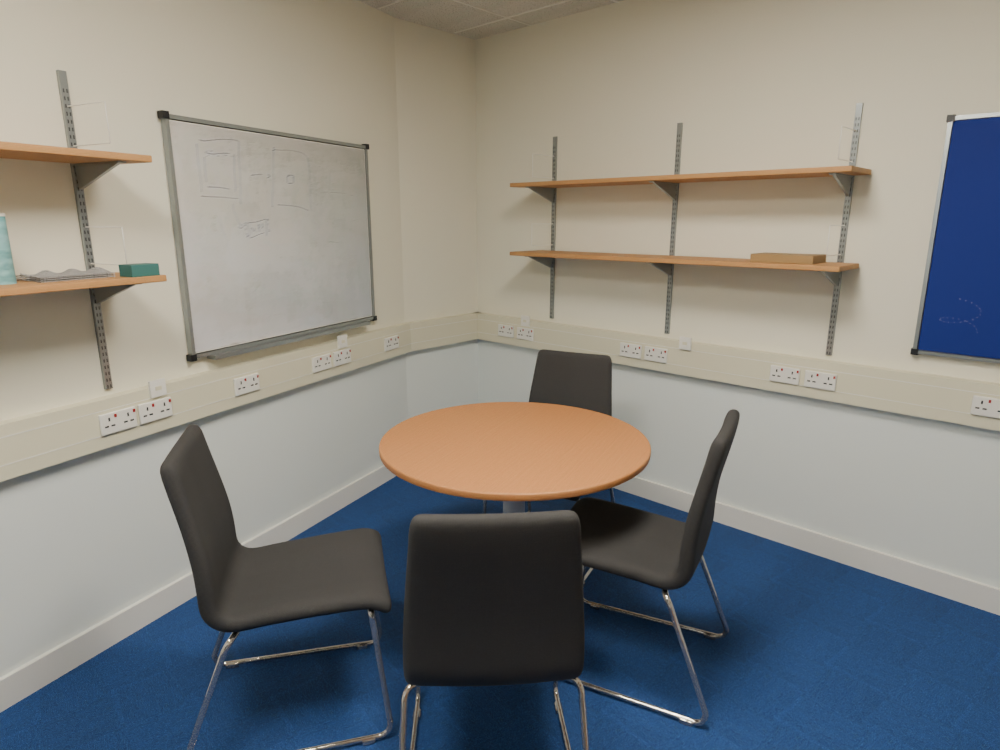}
    \par\scriptsize (a) Original
  \end{minipage}\hfill
  \begin{minipage}[t]{.49\linewidth}
    \centering
    \includegraphics[width=\linewidth]{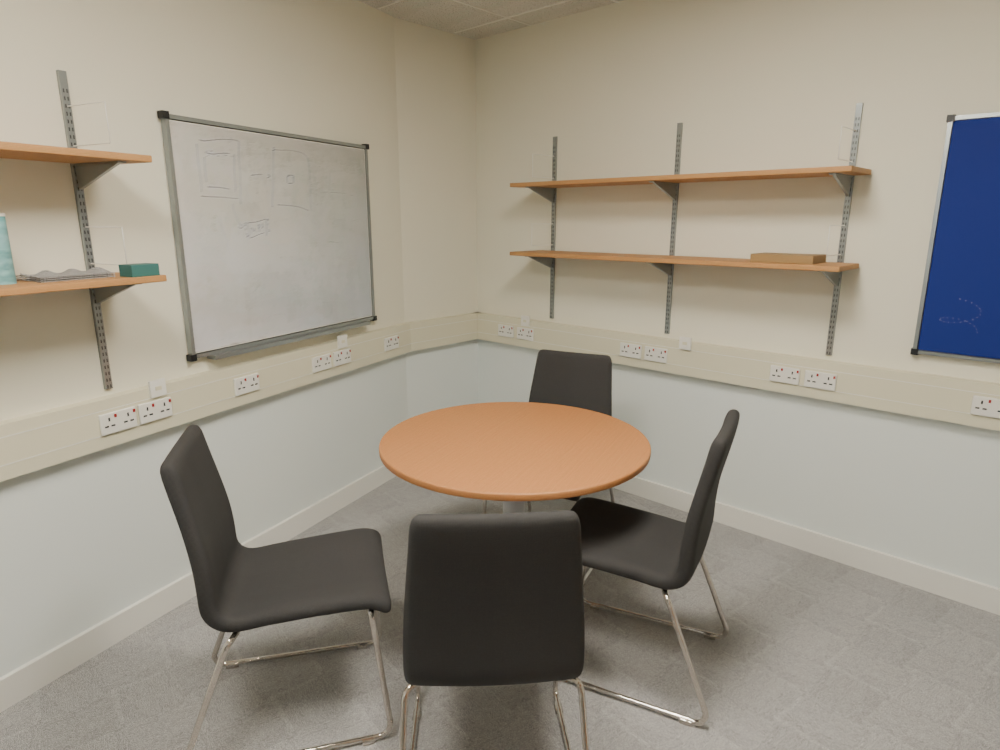}
    \par\scriptsize (b) Grey carpet
  \end{minipage}\par\vspace{4pt}
  \begin{minipage}[t]{.49\linewidth}
    \centering
    \includegraphics[width=\linewidth]{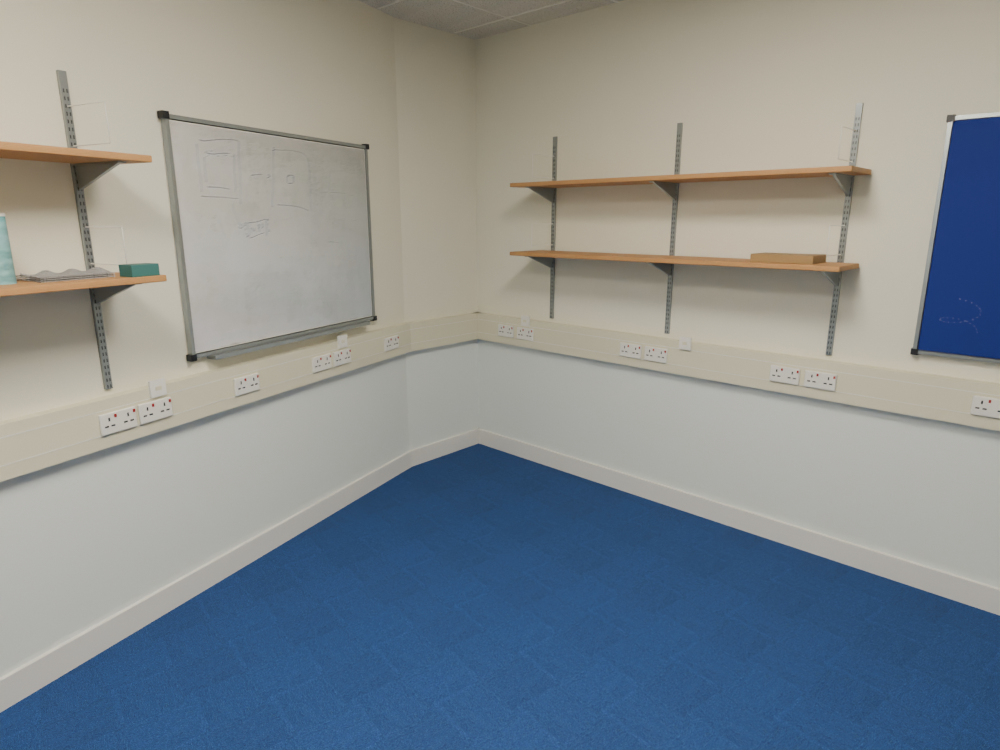}
    \par\scriptsize (c) Cleared area
  \end{minipage}\hfill
  \begin{minipage}[t]{.49\linewidth}
    \centering
    \includegraphics[width=\linewidth]{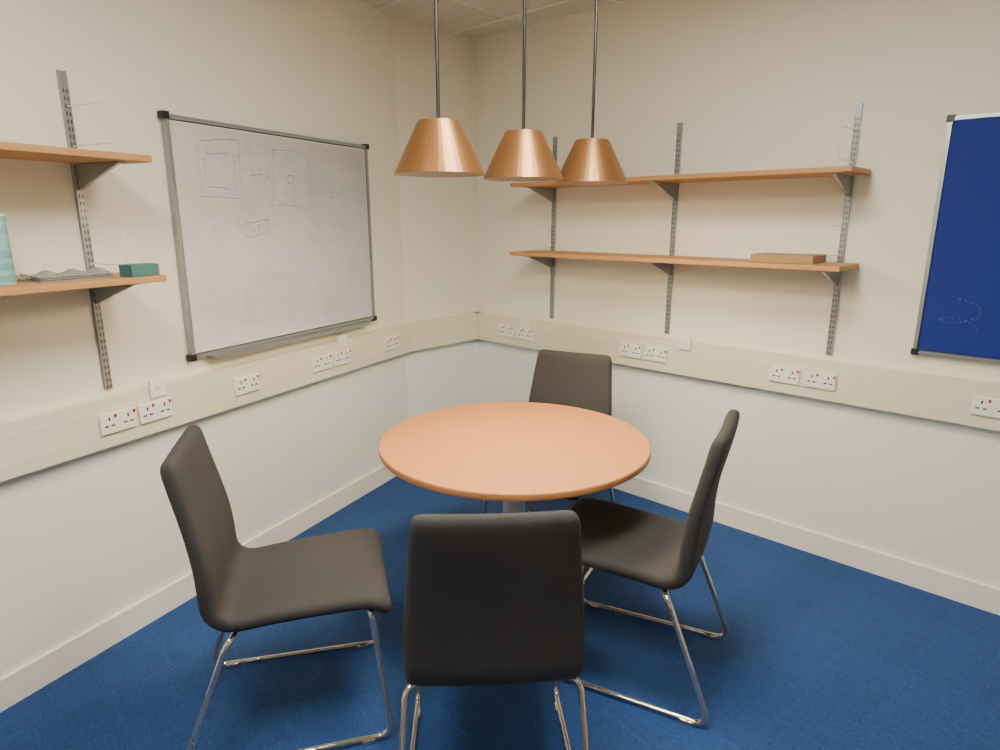}
    \par\scriptsize (d) Pendant lamps
  \end{minipage}
\end{minipage}\hfill
\begin{minipage}[t]{0.485\linewidth}
  \vspace{0pt}
  \centering
  \begin{minipage}[t]{.49\linewidth}
    \centering
    \includegraphics[height=.75\linewidth,viewport=3.05 24.87 196.36 171.27,clip]{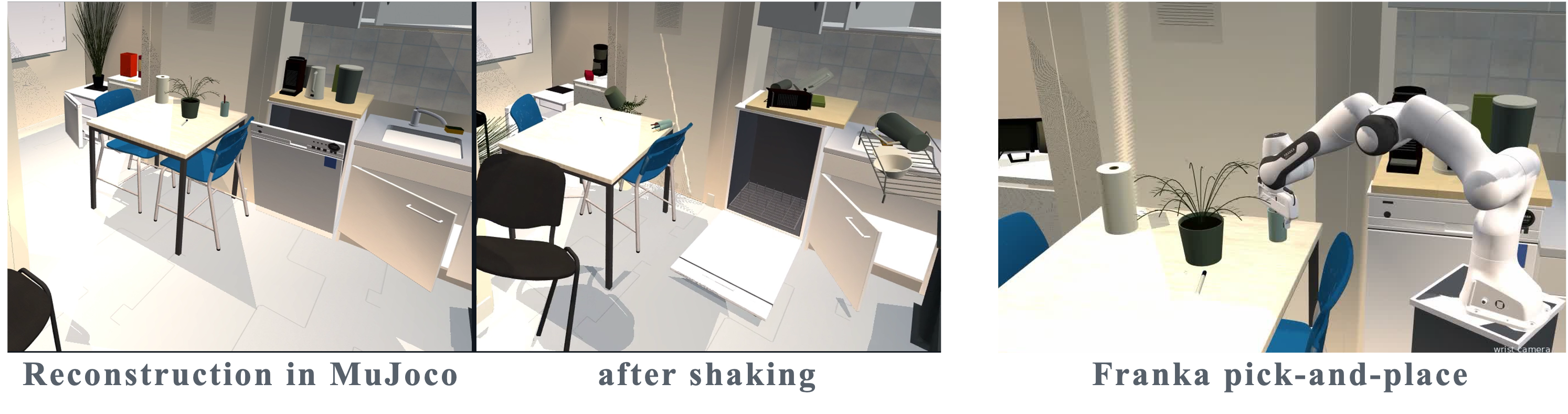}
    \par\scriptsize (a) MuJoCo scene
  \end{minipage}\hfill
  \begin{minipage}[t]{.49\linewidth}
    \centering
    \includegraphics[height=.75\linewidth,viewport=198.55 24.87 391.42 171.27,clip]{figures/main/robotic_application.png}
    \par\scriptsize (b) After shaking
  \end{minipage}\par\vspace{4pt}
  \includegraphics[height=.3675\linewidth,viewport=414.98 24.87 651.27 171.27,clip]{figures/main/robotic_application.png}
  \par\scriptsize (c) Franka pick-and-place
\end{minipage}
\par\vspace{8pt}\nointerlineskip
\begin{minipage}[t]{0.485\linewidth}
  \caption{\textbf{Scene editing.} Programmatic edits to the reconstructed office: carpet recolouring, furniture removal, and pendant-light addition.}
  \label{fig:interaction}
\end{minipage}\hfill
\begin{minipage}[t]{0.485\linewidth}
  \caption{\textbf{Simulation and interaction.} The reconstructed kitchen in MuJoCo, after shaking, and in a Franka pick-and-place example.}
  \label{fig:simulation}
\end{minipage}
\end{minipage}
\end{figure}

Figure~\ref{fig:cases} shows a kitchen and meeting room with complementary capture and reconstruction views, with additional rooms and views provided in Appendix~\ref{app:comparisons}. The \textbf{editable reconstruction} supports changes to materials, furniture, and fixtures without repeating capture or reconstruction; Figure~\ref{fig:interaction} illustrates controlled programmatic edits, while Appendix~\ref{app:editing_demonstrations} provides implementation details and complementary natural-language editing examples. Figure~\ref{fig:simulation} further demonstrates the \textbf{exported kitchen in MuJoCo}, including its response to shaking and a Franka manipulation example. Separate bodies, support relations, and joints enable physical interaction; however, \textbf{these demonstrations do not measure robot-task success or identify real-world dynamics}.

\subsection{Human and Agent-Based Perceptual Evaluation}
\label{sec:perceptual_evaluation}

Human and agent evaluators rate LiteReality~\citep{huang2026litereality}, Fable 5.1 (+Scan),
GPT-6 (+Scan), and LiteReality-Agent on layout accuracy, object similarity,
and overall scene similarity, each on a five-point scale (higher is better).

\paragraph{Evaluation protocol.}
Human participants compare reference images with matched-view
reconstruction renders and orthographic plan views, with method
identities anonymised and positions randomised per scene. Each
participant rates all four methods on five scenes from the ten-scene
benchmark, giving 65 participant--scene evaluations across 13 verified
submissions; ratings are averaged with equal weight and scene coverage
is uneven. Agent-1 and Agent-2 rate the same methods and criteria.
Appendix~\ref{app:perceptual_details} gives the full protocol.

\begin{table}[t]
  \centering
  \input{tables/perceptual_evaluation}
  \caption{Perceptual scores (1--5; higher is better). Human means: 13 submissions, 65 ratings per method. A1 and A2: agent evaluators.}
  \label{tab:perceptual_evaluation}
  \vspace{-3mm}
\end{table}

\paragraph{Results.}
LiteReality-Agent has the highest human means (Table~\ref{tab:perceptual_evaluation}):
layout $4.60$, object similarity $4.58$, and overall similarity $4.54$,
exceeding GPT-6 (+Scan) by $1.49$, $1.60$, and $1.54$, respectively.
For overall similarity, our method wins
53 of 65 paired participant--scene comparisons against GPT-6 (+Scan),
with eleven ties and one loss. Both agent evaluators also rank
LiteReality-Agent first and GPT-6 (+Scan) second; their scores differ
from human ratings. These ratings measure perceived fidelity. The small sample, uneven
coverage, and dependent ratings do not establish statistical significance
or physical validity. Agent ratings do not replace human judgments.

%% file: tables/main_results.tex
\begin{table}[t]
    \centering
    \setlength{\tabcolsep}{3pt}
    \renewcommand{\arraystretch}{1.15}
    \begin{minipage}[b]{0.48\linewidth}
        \centering
        \caption{Mean visual fidelity over 620 capture views from ten
        scenes, under shared cameras and scene-specific lighting.
        Bold is best.}
        \label{tab:visual_results}
        \scriptsize
        \begin{tabular}{@{}lrrrr@{}}
            \toprule[1pt]
            Method & SSIM $\uparrow$ & \shortstack[r]{PSNR\\(dB) $\uparrow$}
                   & RMSE $\downarrow$ & LPIPS $\downarrow$ \\
            \midrule[0.5pt]
            LiteReality              & 0.5609 & 11.881 & 0.2635 & 0.6709 \\
            Fable 5.1 (RGB-only)      & 0.5307 & 10.681 & 0.3008 & 0.7197 \\
            Fable 5.1 (+Scan)         & 0.5649 & 12.287 & 0.2506 & 0.6422 \\
            GPT-6 (RGB-only)          & 0.5742 & 11.549 & 0.2694 & 0.6910 \\
            GPT-6 (+Scan)             & 0.6004 & 12.967 & 0.2323 & 0.6250 \\
            \midrule
            LiteReality-Agent (ours) & \textbf{0.6105} & \textbf{14.129}
                                     & \textbf{0.2033} & \textbf{0.5392} \\
            \bottomrule[1pt]
        \end{tabular}
    \end{minipage}\hfill
    \begin{minipage}[b]{0.48\linewidth}
        \centering
        \caption{Geometric consistency and plausibility over ten scenes.
        Depth MAE over 611 frames; near-contact supported/total.
        Bold is best.}
        \label{tab:layout_evaluation}
        \scriptsize
        \begin{tabular}{@{}lccc@{}}
            \toprule[1pt]
            Method & \shortstack{Depth MAE\\(cm) $\downarrow$}
                   & \shortstack{Bottom\\near-contact $\uparrow$}
                   & \shortstack{Inters.\\pairs $\downarrow$} \\
            \midrule[0.5pt]
            LiteReality         & 15.03 & 85/85 (\textbf{100\%}) & 43 \\
            Fable 5.1 (RGB-only) & 58.64 & 71/102 (69.6\%) & 36 \\
            Fable 5.1 (+Scan)    & 16.32 & 78/97 (80.4\%) & 54 \\
            GPT-6 (RGB-only)     & 40.51 & 65/94 (69.1\%) & 14 \\
            GPT-6 (+Scan)        & 13.33 & 58/92 (63.0\%) & 24 \\
            \midrule
            LiteReality-Agent (ours)
                & \textbf{10.50} & 86/86 (\textbf{100\%}) & \textbf{0} \\
            \bottomrule[1pt]
        \end{tabular}
    \end{minipage}
\end{table}

%% file: tables/perceptual_evaluation.tex
\small
\setlength{\tabcolsep}{3pt}
\renewcommand{\arraystretch}{1.0}
\begin{tabular*}{\linewidth}{@{\extracolsep{\fill}}lccccccccc@{}}
    \toprule[1pt]
    & \multicolumn{3}{c}{Layout $\uparrow$}
    & \multicolumn{3}{c}{Object $\uparrow$}
    & \multicolumn{3}{c}{Overall $\uparrow$} \\
    \cmidrule(lr){2-4} \cmidrule(lr){5-7} \cmidrule(l){8-10}
    Method & Hum. & A1 & A2 & Hum. & A1 & A2 & Hum. & A1 & A2 \\
    \midrule[0.5pt]
    LiteReality          & 2.03 & 3.00 & 3.00 & 1.95 & 1.70 & 1.70 & 2.11 & 1.80 & 1.80 \\
    Fable 5.1 (+Scan)     & 2.80 & 3.10 & 3.00 & 2.66 & 2.50 & 2.30 & 2.82 & 3.00 & 2.30 \\
    GPT-6 (+Scan)         & 3.11 & 4.10 & 3.60 & 2.98 & 3.20 & 2.90 & 3.00 & 3.60 & 2.90 \\
    \midrule
    LiteReality-Agent    & \textbf{4.60} & \textbf{5.00} & \textbf{4.90}
                         & \textbf{4.58} & \textbf{4.80} & \textbf{4.20}
                         & \textbf{4.54} & \textbf{4.90} & \textbf{4.20} \\
    \bottomrule[1pt]
\end{tabular*}

%% file: sections/5_conclusion.tex
\section{Conclusion}
\label{sec:conclusion}

We presented \system{}, an open-source system that treats 3D reconstruction
as a coding problem, turning RGB-D scans into realistic, articulated, and
simulation-ready scenes. Its observe-edit-verify harness combines executable
scene programs, specialised tools, and evidence-driven verification to
improve reconstruction quality and reliability. Across ten scenes, our
results show gains in visual fidelity, geometric consistency, and perceptual
quality over frontier-model agent baselines. As coding agents advance, this
orchestration framework offers a practical foundation for robust real-to-sim
systems and downstream embodied AI.

%% file: sections/appendix/appendix.tex
\section{Additional Reconstruction Comparisons}
\label{app:comparisons}
\begin{figure}[H]
  \centering
  \includegraphics[width=0.85\textwidth]{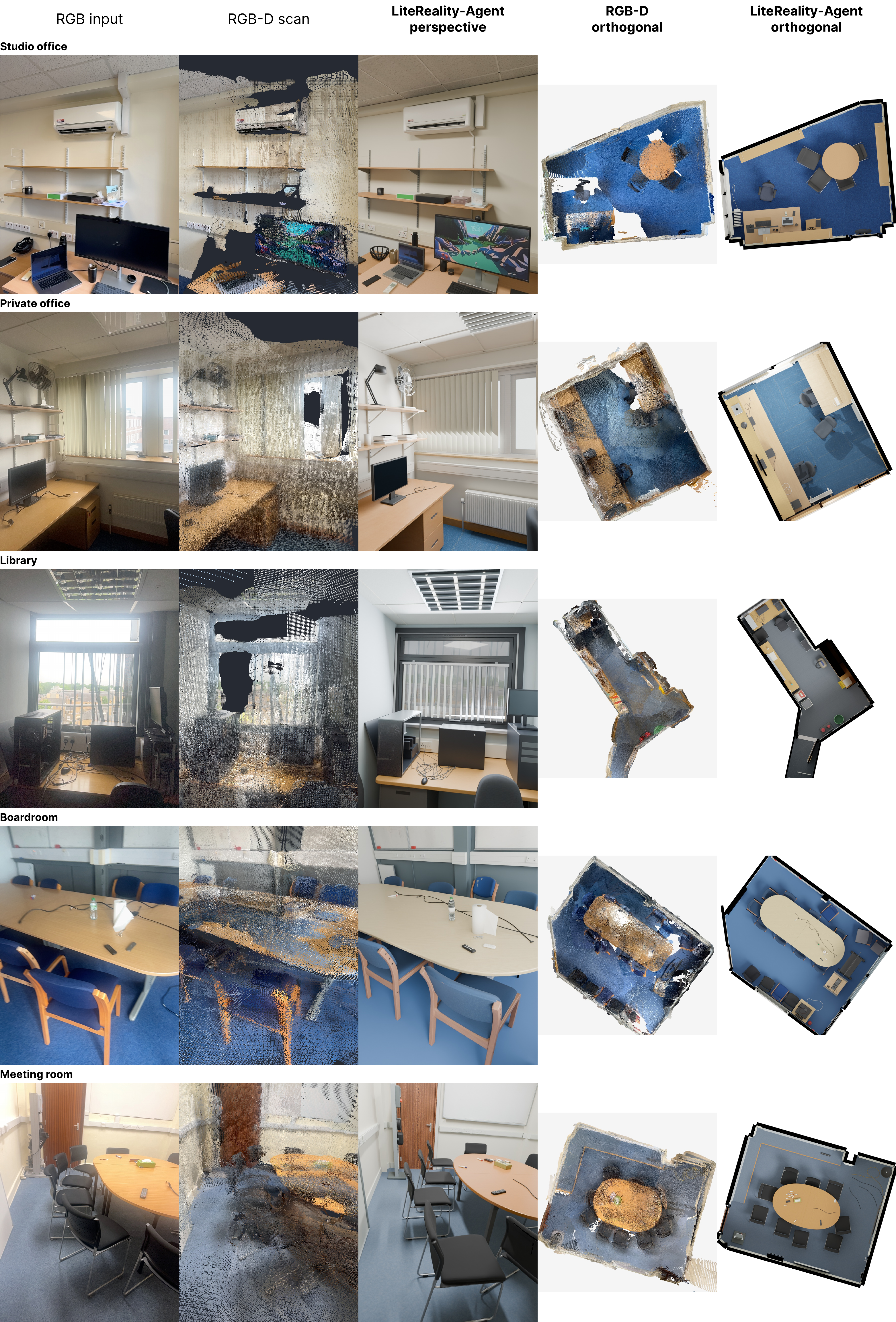}
  \caption{
    Additional complementary view comparisons. Each case shows the RGB
    input, the corresponding RGB--D scan.
  }
  \label{fig:cases-fullgallery1}
\end{figure}

\begin{figure}[H]
  \centering
  \begin{minipage}{.94\linewidth}
    \small
    Figure~\ref{fig:cases-fullgallery2} complements
    Figure~\ref{fig:cases-fullgallery1} with a second matched viewpoint for each room,
    revealing additional scene regions and enabling a broader assessment of
    reconstruction consistency across viewpoints.
  \end{minipage}
  \par\vspace{0.55em}
  \includegraphics[width=0.92\textwidth]{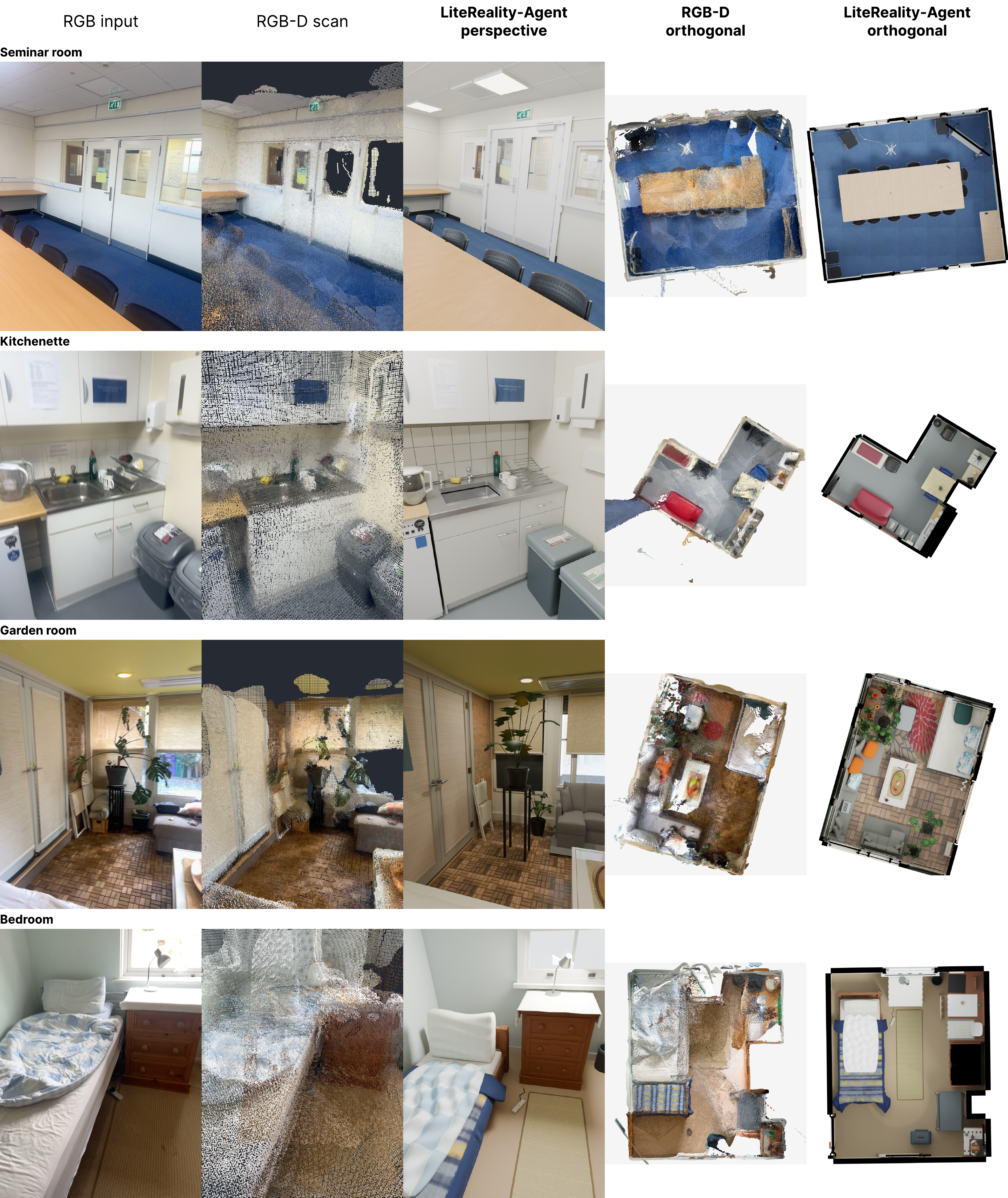}
  \caption{
    \textbf{Complementary reconstruction comparisons across all evaluation rooms.}
  }
  \label{fig:cases-fullgallery2}
\end{figure}

Figures~\ref{fig:cases-fullgallery1} and~\ref{fig:cases-fullgallery2}
show all nine reconstructed rooms using the same camera calibration and
image crop as the main gallery, pairing human-height views with
orthographic plan views.

\clearpage
\clearpage

\input{sections/appendix/system_details}
\clearpage

\section{Additional Experimental Details}
\label{app:experimental_details}

\paragraph{Scan inputs.}
For each scene, the +Scan condition supplements the RGB photographs
with aligned LiDAR depth and confidence maps, camera intrinsics,
poses, timestamps, a derived point cloud, and a RoomPlan USDZ layout
collected or generated by LiteReality Scan.
The RoomPlan layout encodes room structure and the semantic
categories, dimensions, positions, and orientations of detected
objects. LiteReality and LiteReality-Agent receive the same scan
information as the +Scan baselines.

\paragraph{Evaluation views and camera alignment.}
The ten evaluation scenes contain 27--91 capture views each.
All six method--input configurations are evaluated on the same
scene--frame pairs, yielding 3,720 reference--render comparisons.
Reconstructed Blender scenes are aligned to the coordinate system
of the recorded camera poses while preserving scene scale and
individual object placements. Rendering uses the recorded camera
poses, with intrinsics adjusted to $960\times720$.

\paragraph{Rendering configuration.}
All scenes are rendered in Blender 5.2 using Cycles, with up to
64 samples per pixel and OptiX denoising.
For each scene, all methods use the same light objects, world
illumination, and colour-management settings from the corresponding
LiteReality-Agent scene, while retaining their own materials,
textures, and material emission.
Architecture hidden for cutaway views is restored where identified.
The shared lighting is not independently calibrated to the
reference photographs.

\paragraph{Image preprocessing.}
Reference photographs are downsampled from $1920\times1440$ to
$960\times720$ using Lanczos interpolation and paired with renders
by scene and frame identifier.
No additional warping, cropping, masking, rotation, or exposure/colour
correction is applied. All capture frames, including dark renders,
are retained.

\paragraph{Metric implementation.}
SSIM, PSNR, and RMSE are computed on stored RGB values normalised
to $[0,1]$, without linearisation.
SSIM uses an $11\times11$ Gaussian window with $\sigma=1.5$,
population covariance, and averaging across RGB channels.
LPIPS uses pretrained AlexNet with version-0.1 calibration weights,
evaluated at $960\times720$ with inputs normalised to $[-1,1]$.

\paragraph{Aggregation.}
For each metric, we report the arithmetic mean of per-frame scores
over all 620 views. In particular, PSNR is averaged over individually
computed per-frame PSNR values.
Views receive equal weight, so scenes with more capture views
contribute more to the aggregate.

\section{Geometric Evaluation Details}
\label{app:geometric_details}

\paragraph{Depth agreement.}
We compute the mean absolute error between rendered and captured
depth using pixels with valid depth in the capture and all
evaluated outputs, including RoomPlan.
We average per-frame errors over the 611 frames with shared
valid pixels and report depth MAE in centimeters.

\paragraph{Bottom near-contact.}
We count furniture objects with at least one sampled bottom
point within $1$\,cm of an external supporting surface.
We report both the count and its percentage relative to all
identified furniture objects in each method's output.
This criterion measures bottom support proximity; it does not
require every leg or base component to contact a supporting
surface.

\paragraph{Furniture intersections.}
We detect strictly intersecting triangle surfaces in the
evaluated source meshes using a $0.1$\,mm plane-side tolerance.
This tolerance is not a penetration-depth threshold.
Each furniture pair is counted once per scene, and counts
are summed across the ten scenes.
Intersections within a single furniture object and between
furniture and architecture are excluded.
The resulting counts describe detected surface intersections
and require semantic interpretation.

\paragraph{Object populations.}
Object-based metrics use the furniture objects identified
in each method's output. These populations differ across
methods, so near-contact rates and intersection counts
must be interpreted alongside the reported furniture counts.

\section{Additional Perceptual Evaluation Details}
\label{app:perceptual_details}

\paragraph{Rating criteria.}
We evaluate LiteReality, Fable 5.1 (+Scan), GPT-6 (+Scan), and
LiteReality-Agent on three criteria using a five-point scale,
with higher scores indicating closer agreement with the reference
scene. Layout accuracy concerns room structure and object
positions and orientations. Object similarity concerns object
shape, scale, and appearance. Overall scene similarity assesses
the reconstruction as a whole.
We report the criteria separately without constructing a
composite score.

\paragraph{Questionnaire design.}
The questionnaire presents a reference RGB image alongside
reconstruction renders from the corresponding camera viewpoint,
together with an orthographic plan-view comparison.
Methods are displayed anonymously, with their positions
randomised independently for each scene; displayed letters
therefore do not identify a fixed method across scenes.
Each participant evaluates five scenes sampled without
replacement from the ten-scene benchmark and rates all four
methods on all three criteria.

\paragraph{Sample inclusion and aggregation.}
The reported human results use 13 completed submissions with
resolved method identities, comprising 65 participant--scene
evaluations and 780 individual ratings.
Each method receives 65 ratings per criterion.
We report arithmetic means, giving each participant--scene
evaluation equal weight. Scene coverage ranges from three to
eleven participants, so scenes are not equally weighted.
Three additional submissions containing unresolved anonymous
method identifiers are excluded from the current aggregation
pending verification of the identifier mapping.

\paragraph{Agent-based evaluation.}
We report two sets of agent-generated ratings, designated
Agent-1 and Agent-2, for the same four methods and three criteria.
Agent ratings are aggregated and presented separately from
human ratings.

\paragraph{Additional comparisons.}
LiteReality-Agent achieves the highest human mean overall
scene similarity in each of the ten scenes.
Its mean advantage over GPT-6 (+Scan) on this criterion is
$1.54$ points for humans and $1.30$ points for each agent evaluator.
Agent-1 scores LiteReality-Agent above the human mean on all
three criteria, whereas Agent-2 assigns a higher layout score
but lower object and overall similarity scores.
Thus, the evaluations agree on the two leading methods while
differing in absolute score levels.

\clearpage
\input{sections/appendix/limitations}

%% file: sections/appendix/system_details.tex
\section{LiteReality-Agent Implementation Details}
\label{app:system_details}
This appendix supplements Sec.~\ref{sec:litereality-agent} with the capture
format, implementation choices, and intermediate visual examples.

\subsection{Capture Data and Layout Repair}
\label{app:capture_layout}
Table~\ref{tab:app_capture} details the scan data used for reconstruction.

\begin{table}[h]
  \centering
  \small
  \begin{tabular}{@{}p{0.27\linewidth}p{0.68\linewidth}@{}}
  \toprule
  Capture component & Role in reconstruction \\
  \midrule
  RGB frames & Appearance references for objects, materials, and fixtures. \\
  Depth and confidence & Geometric evidence and visibility checks. \\
  Camera calibration & Intrinsics and poses for projection and matched-view rendering; timestamps associate observations. \\
  Derived point cloud & A spatial view of the captured surfaces and missing regions. \\
  RoomPlan detections & Room structure, openings, and semantic object boxes with metric dimensions and poses. \\
  \bottomrule
  \end{tabular}
  \caption{Capture data used by LiteReality-Agent. The point cloud is
  derived from RGB--D observations; the detections provide initial layout
  estimates rather than final object geometry.}
  \label{tab:app_capture}
\end{table}

\paragraph{Graph construction and support.}
Graph nodes represent the room, structural surfaces, openings, and objects.
Relations record opening hosts, support, wall adjacency, and functional
groups. Height and footprint overlap identify support candidates, with
wall-hung objects distinguished from floor-standing furniture. Objects
without plausible support are flagged. Expected containment, including
chairs beneath tables, is distinct from conflicting footprints.

\begin{figure}[h]
  \centering
  \includegraphics[width=.40\linewidth]{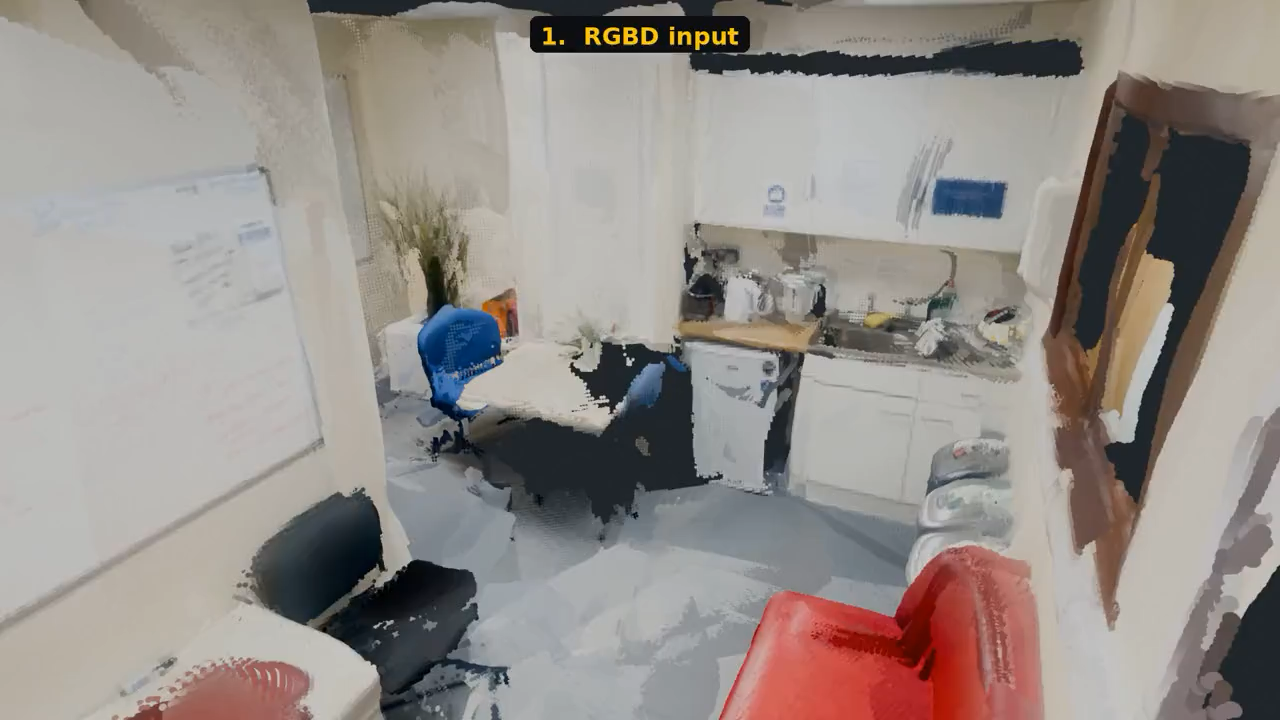}\hspace{.02\linewidth}
  \includegraphics[width=.40\linewidth]{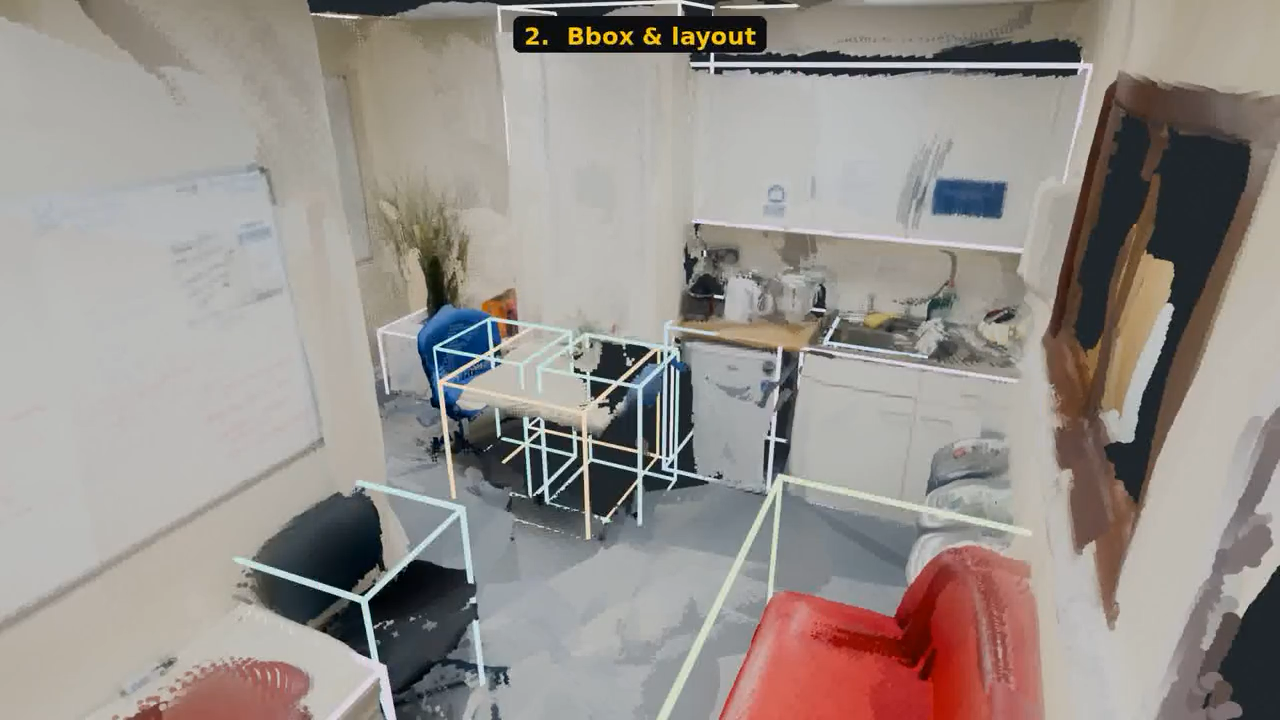}\\[3pt]
  \includegraphics[width=.40\linewidth]{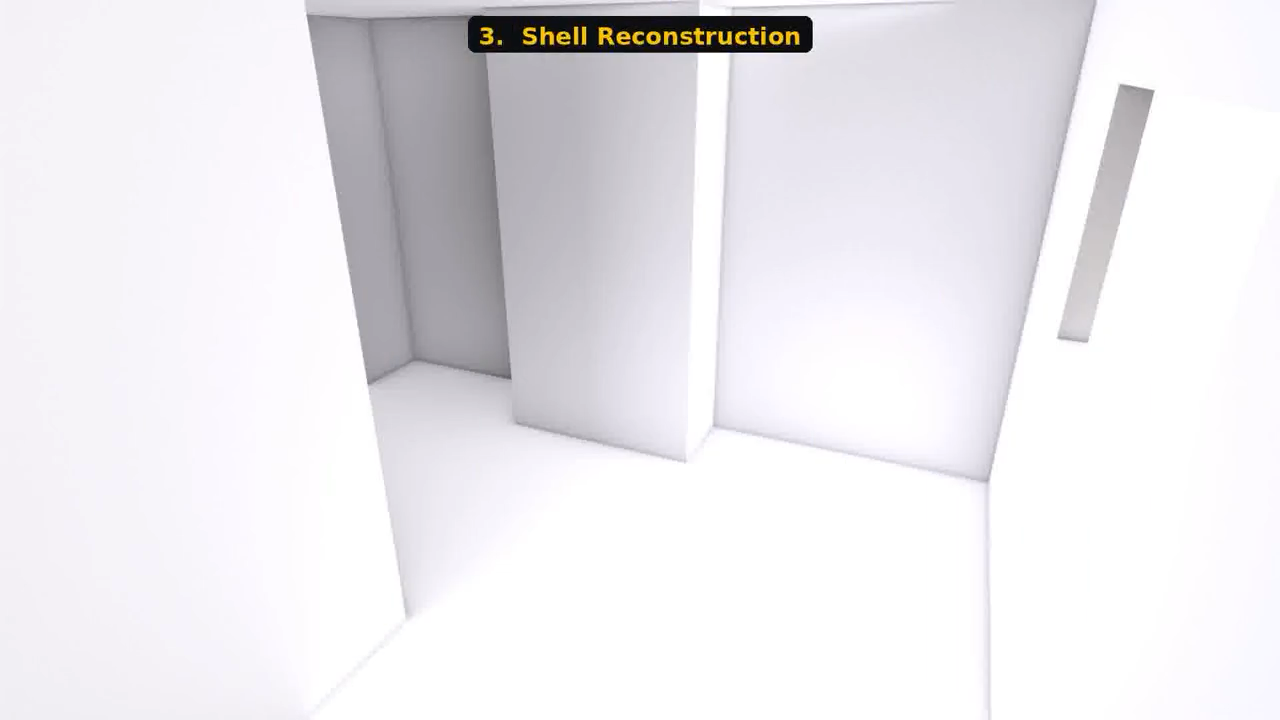}\hspace{.02\linewidth}
  \includegraphics[width=.40\linewidth]{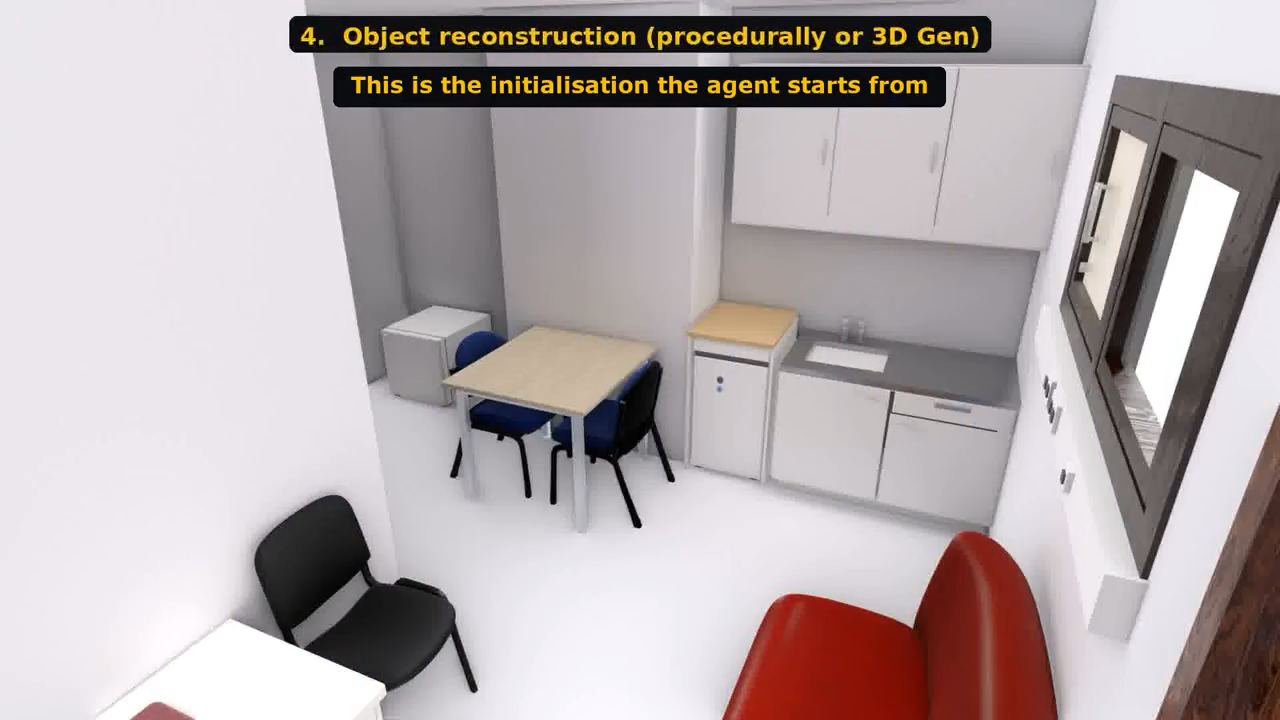}
  \caption{\textbf{From capture to an initial room.} The running kitchen
  example shows the RGB--D point cloud, detected boxes and layout, the
  reconstructed shell, and the assembled initial scene. Fine surface
  appearance, fixtures, and small objects are refined during authoring.}
  \label{fig:app_initialisation}
\end{figure}

\paragraph{Bounded repair.}
Before cropping and generation, candidate layouts are compared by error
count, then summed error magnitude. Fidelity limits apply cumulatively
against the original scan. The implementation caps horizontal displacement at
$0.6$\,m and vertical displacement at $0.08$\,m. Per-axis size changes are
limited to $15\%$ for tables, desks, chairs, sofas, and beds, and $35\%$
for other categories. Erroneous merges may be split into their original
members; measured objects cannot be deleted to remove violations.
Repair stops on validation, repeated lack of progress, or six agent-assisted
rounds. Corrected boxes are handed to room assembly
(Fig.~\ref{fig:app_initialisation}).

\clearpage
\subsection{Layout Agent: Relations, Repair, and Before--After Results}
\label{app:layout_agent}
The layout stage corrects the detected boxes before they determine object
crops, reconstruction dimensions, and placement. It combines a geometric
scene graph, a deterministic repair search, and an optional image-guided
agent for unresolved cases. This ordering avoids propagating a misplaced
or oversized detection into asset generation.

\paragraph{Relations constrain the edits.}
The graph distinguishes physical support from functional grouping.
An object has one support parent---a floor, wall, ceiling, or another
object---or is flagged as unsupported. A chair may belong to a desk's
functional group while being supported by the floor. This distinction
allows a desk and its chair to move together without claiming that the
desk supports the chair. Separate relations encode wall adjacency,
opening hosts, and expected containment, such as a chair tucked under
a table. These category-aware exceptions prevent plausible box overlaps
from being treated as errors.

\paragraph{Propose, validate, and retain.}
The geometric search tries wall alignment, corner fitting, separation,
and small changes to box dimensions. Each candidate is checked against
the room boundary, walls, other objects, and the scan-fidelity limits
above. When geometry alone leaves an unresolved case, the optional
agent receives the current box, nearby walls and objects, the violation,
reference photographs, and feedback from rejected proposals. It can
propose translation, resizing, attachment to one or two walls, or
splitting a merged box into its original members. Proposals must preserve
existing anchors, satisfy cumulative fidelity bounds, and improve the
geometric score before they are retained. Rejected edits leave the
accepted layout intact; unresolved findings remain in the report.

\paragraph{Recorded outcomes and scope.}
The saved runs reduce the layout checker's error count from $30$ to $0$.
Eight scenes required changes; the
two initially valid scenes were unchanged. Across $107$ input boxes,
$29$ were moved and four were resized. These runs were resolved by the
deterministic search and therefore do not measure the additional benefit
of image-guided proposals. A zero count means that the boxes pass this
checker's tolerances and category exemptions. Generated meshes and later
authored objects still require the separate collision and support checks
in Appendix~\ref{app:qc_simulation}.

\clearpage
\begin{figure}[t]
  \centering
  \includegraphics[width=\linewidth]{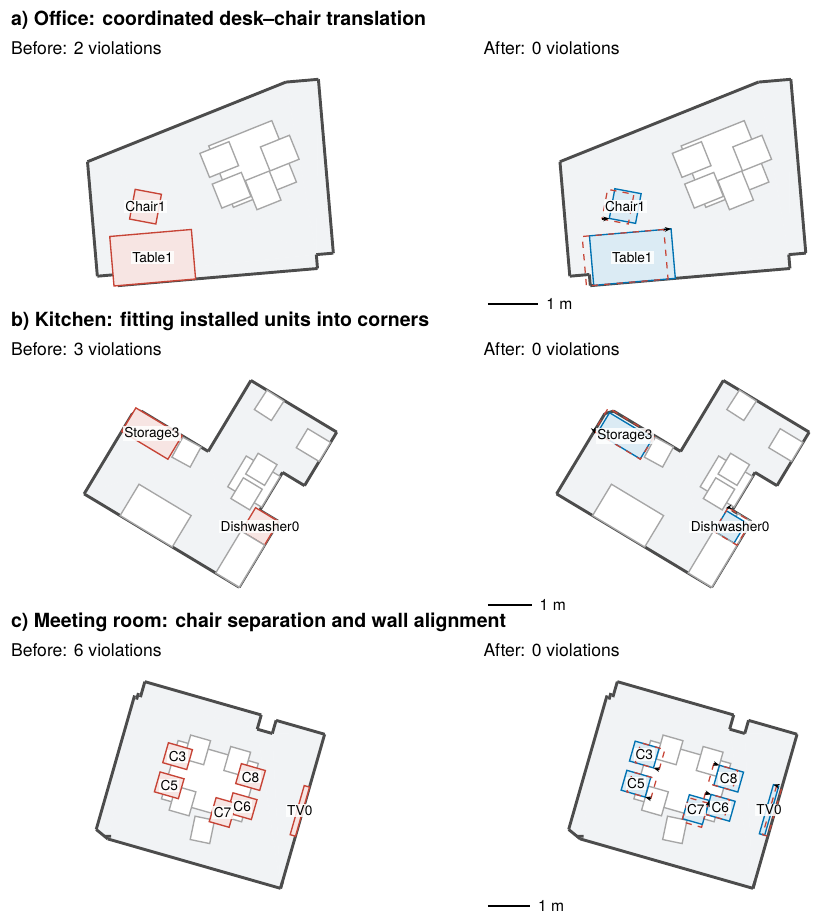}
  \caption{\textbf{Before and after layout repair.}
  Plans are drawn directly from the saved input and output boxes for
  three scenes. Red outlines in the left column identify boxes that
  are changed; blue outlines show their corrected states on the right.
  Dashed red outlines retain their original footprints, and arrows show
  centre translations. Each pair uses the same scale and viewpoint;
  grey objects and walls provide unchanged context. Meeting-room labels
  abbreviate chairs as C and the television as TV. These are layout-stage
  comparisons, before detailed mesh reconstruction and authoring.}
  \label{fig:app_layout_before_after}
\end{figure}

\paragraph{Coordinated movement.}
In the office (Fig.~\ref{fig:app_layout_before_after}a),
\texttt{Table1} and \texttt{Chair1} both move by $(0.1445,0.0124,0)$\,m.
The shared $14.5$\,cm translation preserves their relative arrangement
and removes the two recorded violations without resizing either box.

\paragraph{Corner fitting and local separation.}
In the kitchen (b), \texttt{Storage3} and \texttt{Dishwasher0} each move
about $5.3$\,cm and reduce their two horizontal dimensions by about
$8.5$\,cm, resolving three boundary/wall violations. In the meeting room
(c), five chairs are separated and the television is aligned to its wall,
reducing six violations to zero. All six boxes retain their dimensions;
the largest centre shift is $14.1$\,cm.

\clearpage
\subsection{Object Reconstruction and Physical Metadata}
\label{app:object_details}
\begin{figure}[h]
  \centering
  \includegraphics[width=.80\linewidth]{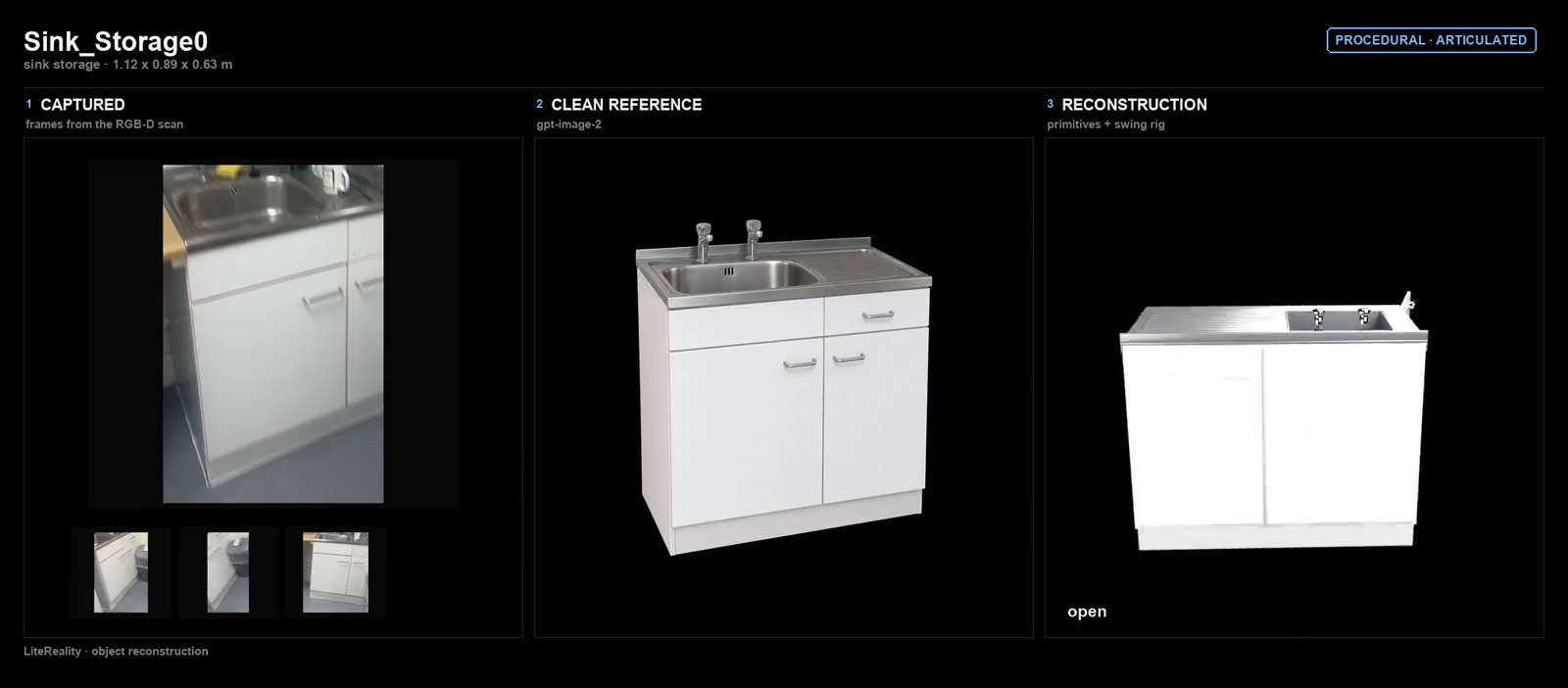}\\[4pt]
  \includegraphics[width=.80\linewidth]{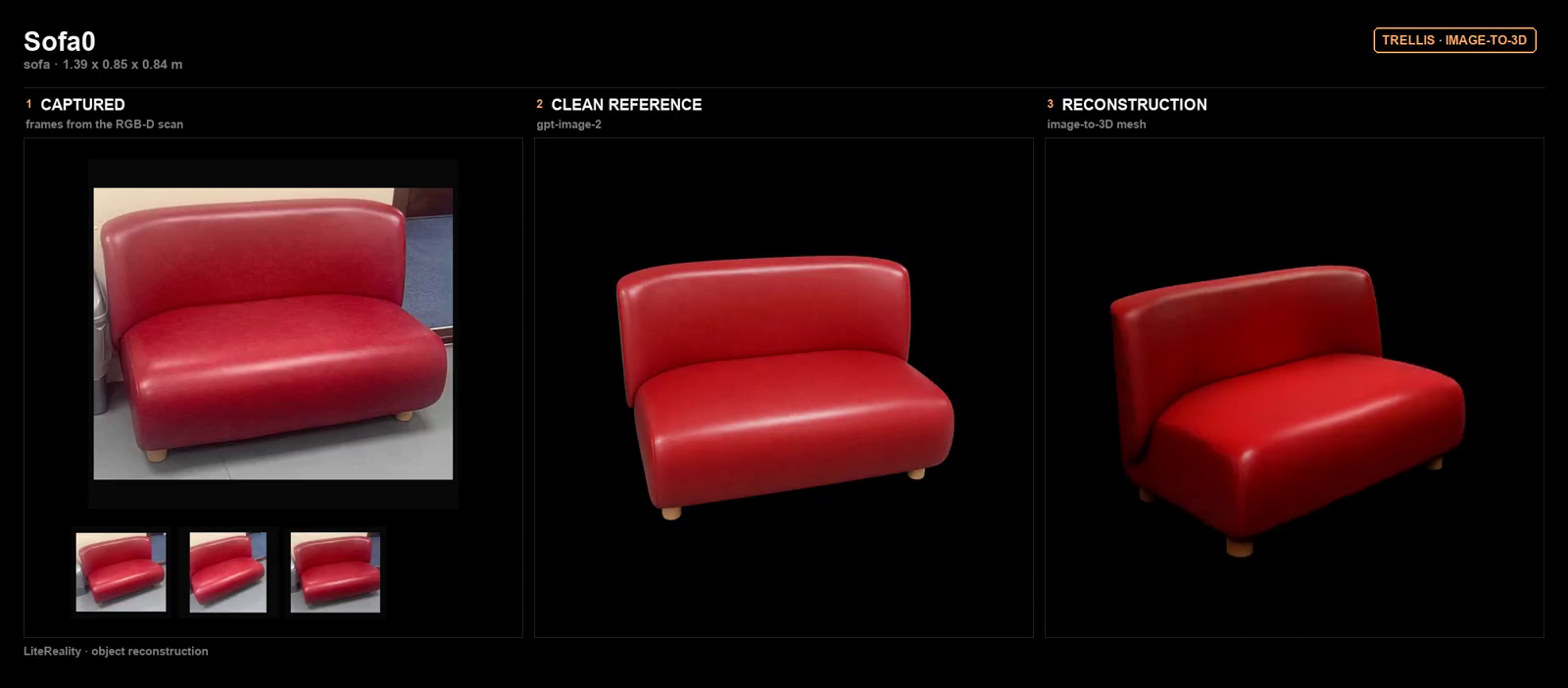}
  \caption{\textbf{Object reconstruction from capture references.}
  Each row shows captured views, the cleaned reference, and a reconstructed
  asset. The sink cabinet uses procedural geometry with articulated parts;
  the sofa uses image-to-3D generation.}
  \label{fig:app_object_routes}
\end{figure}

Figure~\ref{fig:app_object_routes} shows both reconstruction routes using
clean references generated from scan views. Raw captures remain the
evidence for checking fidelity: completed occluded parts are estimates.

\paragraph{Generated-mesh rejection and retries.}
Mesh checks look for implausible aspect ratios, disconnected fragments,
extensive flat boundary surfaces, and dense bottom geometry suggesting
that a floor or background was fused into the asset. The chair repair
loop regenerates both the image reference and the mesh, varying the
reconstruction seed between attempts. A retry is checked again before
acceptance. If the retry budget is exhausted, the implementation retains
the original asset and records the unresolved failure.

\paragraph{Estimating physical properties.}
The procedural program defines links, joint axes, pivots, and limits.
Authored mass or density overrides automatic estimates. Otherwise,
category-dependent occupancy density times bounding-box volume estimates
total mass, which is distributed across links using geometric volume
estimates. Inertia and centre of mass are computed from link geometry;
convex-hull and box estimates provide fallbacks for degenerate meshes.
Material-name priors supply friction and restitution when unspecified.
Restitution is retained in the asset metadata but is not directly used
by the MJCF exporter. Each physical-property record retains its source.

\paragraph{Collision geometry and solver checks.}
Near-convex parts can use a single hull; larger concave parts use convex
decomposition. The decomposition budget depends on link mass to limit
excessive contact complexity for lightweight objects. Per-object MuJoCo
checks comprise a short drop, individual joint release, and tilted-gravity
sliding. They report penetration, separation, unstable motion, and
friction behaviour. Joint tests distinguish frame collisions from sibling
interactions that may require a particular opening sequence.

\clearpage
\subsection{Evidence Tools and Authoring State}
\label{app:evidence_tools}
\begin{figure}[h]
  \centering
  \includegraphics[width=.93\linewidth]{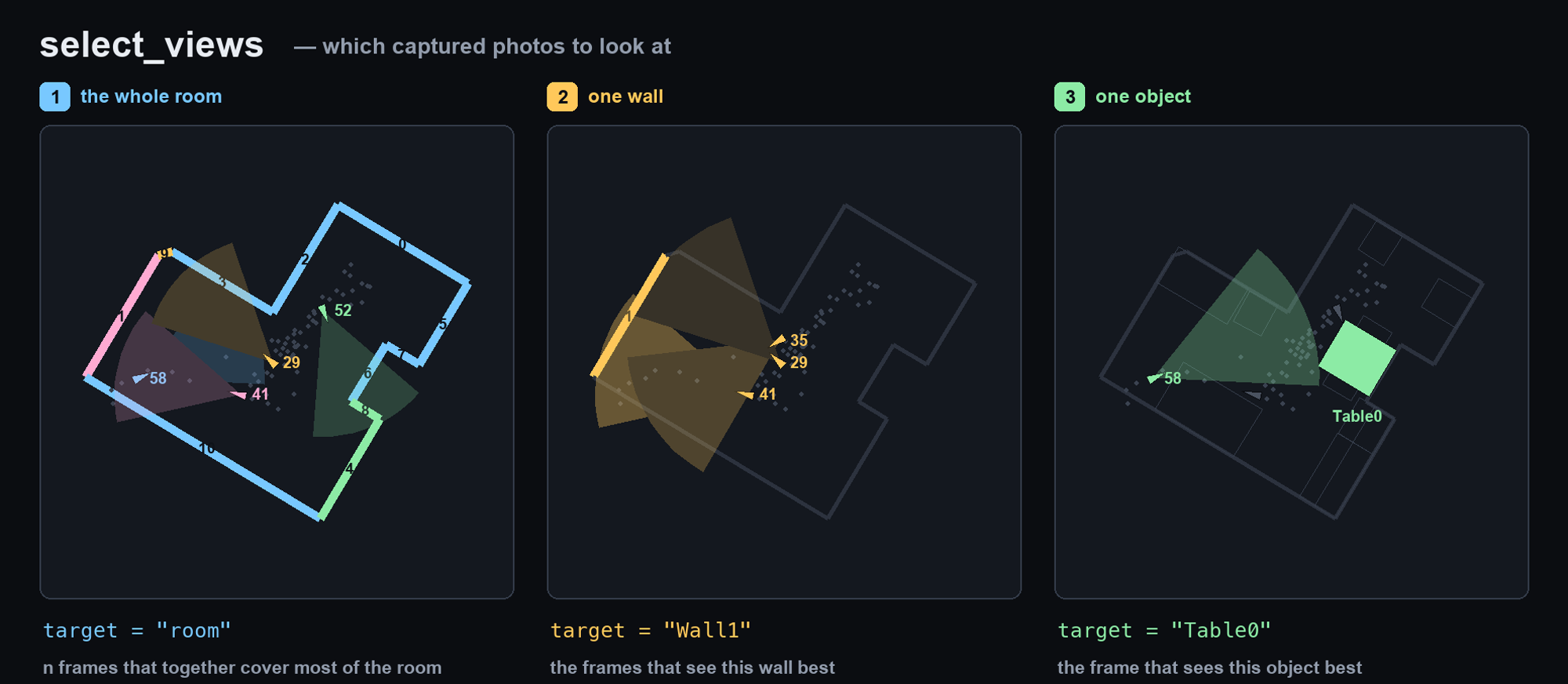}
  \caption{\textbf{View selection at three scopes.} Room queries cover
  several surfaces, wall queries focus on one structural element, and
  object queries return observations of an individual asset. Numbers
  identify capture frames.}
  \label{fig:app_view_selection}
\end{figure}

\paragraph{Selecting and comparing views.}
The view selector supports room, wall, and object queries
(Fig.~\ref{fig:app_view_selection}). Room queries seek complementary
coverage; local queries rank views of the requested target. The render
tool rebuilds the current program and pairs the output with its capture
photograph using recorded calibration. This exposes discrepancies in
shape, placement, and appearance without asking the agent to align
unrelated viewpoints.

\begin{figure}[h]
  \centering
  \begin{minipage}[b]{.31\linewidth}
    \includegraphics[width=\linewidth]{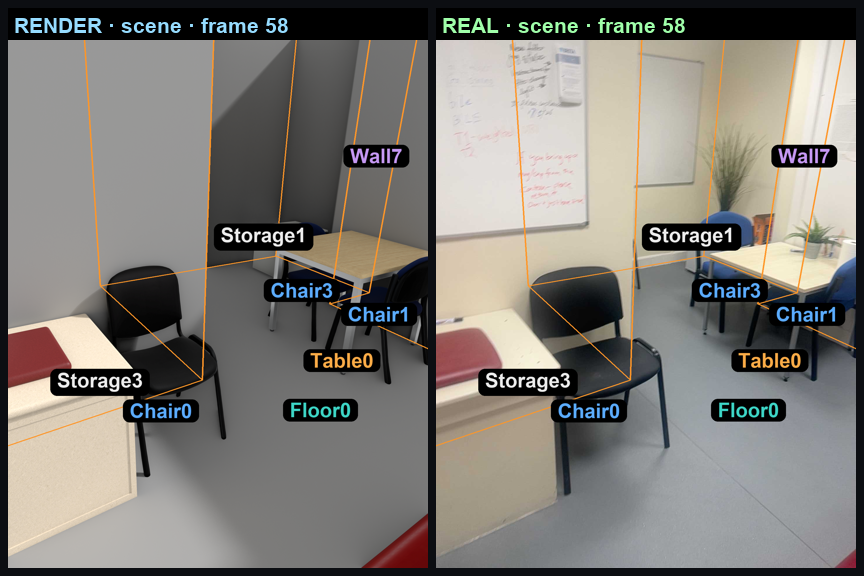}
    \centering\scriptsize (a) Room: object labels
  \end{minipage}\hfill
  \begin{minipage}[b]{.31\linewidth}
    \includegraphics[width=\linewidth]{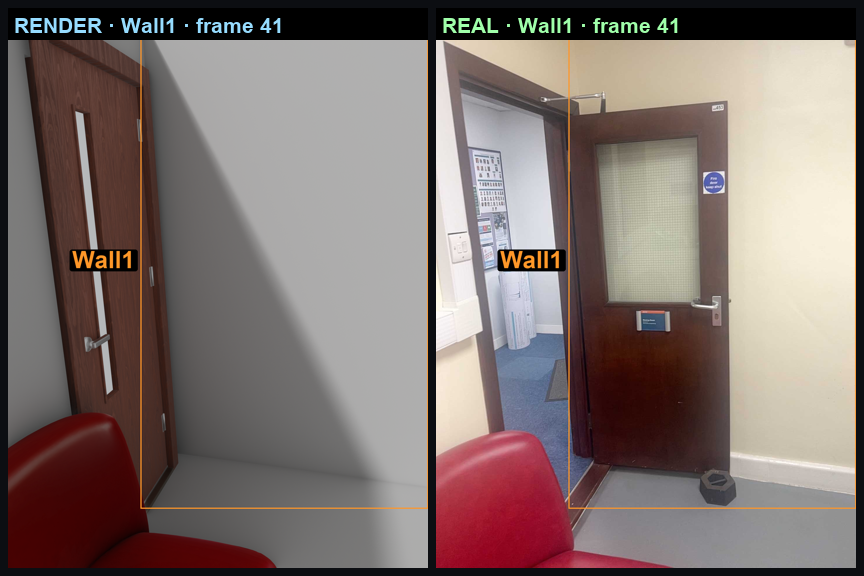}
    \centering\scriptsize (b) Wall: selected outline
  \end{minipage}\hfill
  \begin{minipage}[b]{.31\linewidth}
    \includegraphics[width=\linewidth]{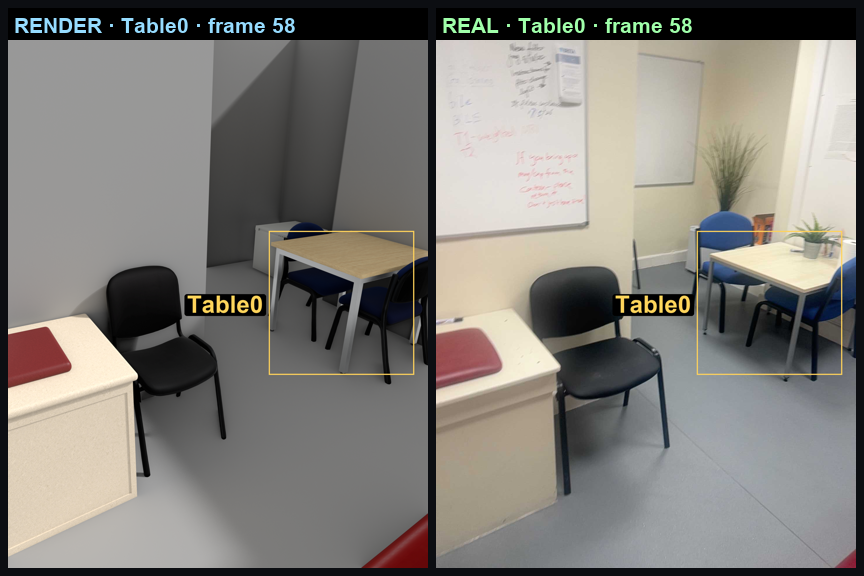}
    \centering\scriptsize (c) Object: bounding box
  \end{minipage}
  \caption{\textbf{Evidence for authoring.} (a--c) Render--photograph
  pairs use room, wall, and object annotations to locate discrepancies
  at progressively finer scopes.}
  \label{fig:authoring-evidence}
\end{figure}

\paragraph{Persistent editing and materials.}
\texttt{Room.py} contains room-level construction and placement, while
\texttt{object.py} contains a procedural asset's build recipe. Material
retrieval saves colour, roughness, and normal maps together with a reusable
recipe; recolouring changes the diffuse appearance while retaining the
spatial pattern. Rebuilding these programs is the handoff between
modules, ensuring that the exported geometry reflects the saved edits.

\clearpage
\subsection{Stitched Wall Visualizations}
\label{app:wall_stitching}
\begin{figure}[h]
  \centering
  \begin{minipage}[b]{.48\linewidth}
    \includegraphics[width=\linewidth]{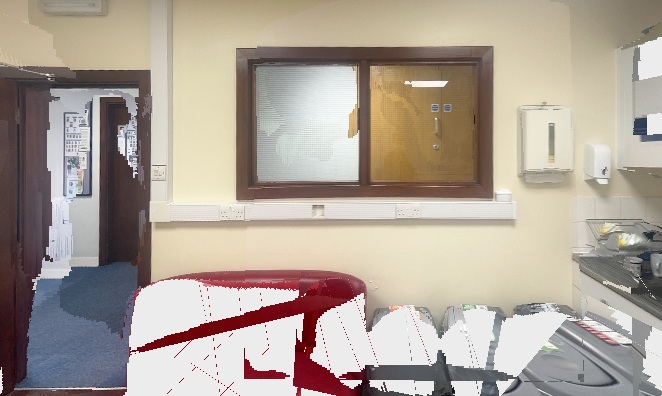}
    \centering\small (a) Stitched wall reference
  \end{minipage}\hfill
  \begin{minipage}[b]{.48\linewidth}
    \includegraphics[width=\linewidth]{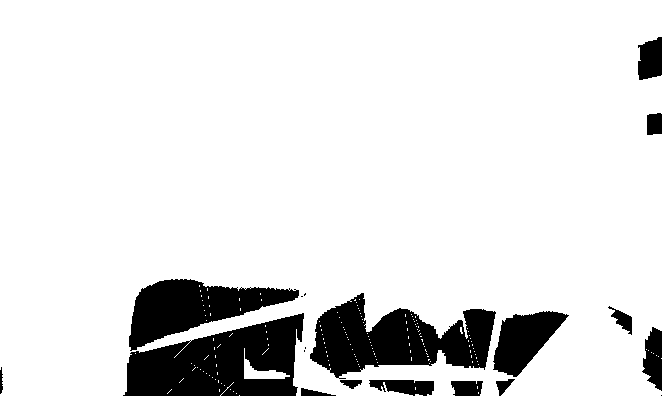}
    \centering\small (b) Observed-pixel coverage mask
  \end{minipage}
  \par\vspace{10pt}
  \begin{minipage}[b]{.36\linewidth}
    \centering
    \includegraphics[width=.84\linewidth]{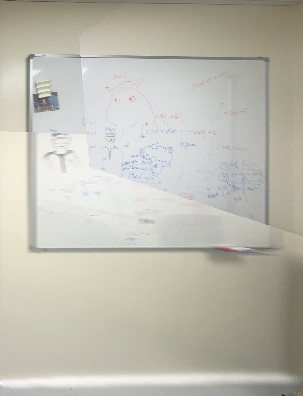}
    \par\small (c) Stitched whiteboard wall
  \end{minipage}\hspace{.06\linewidth}
  \begin{minipage}[b]{.36\linewidth}
    \centering
    \includegraphics[width=.84\linewidth]{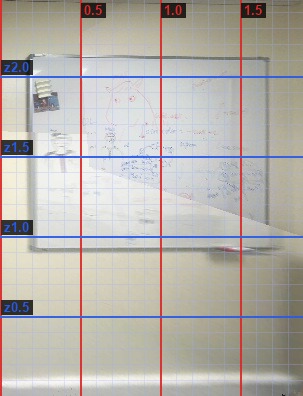}
    \par\small (d) Metric-grid overlay
  \end{minipage}
  \caption{\textbf{Wall stitching and metric measurement.}
  Two walls from the running kitchen example are rectified from calibrated
  capture frames. (a) The stitched reference retains visible seams and
  projection artefacts around protruding furniture. (b) Its coverage mask
  distinguishes observed pixels (white) from unobserved pixels (black).
  (c--d) A second stitched wall is shown before and after adding a metric
  grid: red ticks measure distance along the wall and blue ticks measure
  height above the floor, in metres. The whiteboard edges can be read
  against this grid to guide placement and sizing in the scene program.}
  \label{fig:app_wall_stitching}
\end{figure}

\paragraph{Wall stitching and measurement.}
Figure~\ref{fig:app_wall_stitching} shows stitched wall references,
their coverage, and the metric overlay. A metric grid is laid on the detected wall plane. Each output pixel
corresponds to a 3D location that is projected into candidate frames using
their calibration. Selection favours frontal, central, nearby observations,
with depth visibility used when available. Selected frames fill the
rectified image without blending; a coverage mask records unobserved
pixels. The metric-grid tool overlays distance along the wall and height
above the floor. These coordinates can be used directly to place
fixtures. The planar projection can distort protruding objects, so raw
capture views remain necessary for lamps, clutter, and furniture.

\clearpage
\subsection{Quality Control and MuJoCo Integration}
\label{app:qc_simulation}
\paragraph{Collision and support are separate checks.}
A chair can overlap a table's box without intersecting its actual legs or
top. We therefore use mesh intersections to establish clashes and convex
approximations to estimate a correction. The resolver moves unanchored
objects horizontally within a bounded correction budget, writes the
accumulated changes to the scene program, and rebuilds before rechecking.
It reports cases that cannot be solved by these local motions.

For a declared support, the validator compares the object's underside
with the support's top surface and measures footprint overlap. Gaps
above $1$\,cm and penetration exceeding $2$\,cm are blocking findings;
less than half of the footprint overlapping the support is reported for
review. Missing or unknown support declarations also block validation.
For recognised floor-standing objects without an explicit declaration,
the floor-contact fallback uses a $5$\,cm tolerance. Structural attachments
are recorded separately and are exempt from gravity-support checks.
These implementation checks differ from the evaluation metric in
Appendix~\ref{app:geometric_details}, which measures sampled bottom
near-contact and is not the authoring gate.

\paragraph{Visual review and bounded sessions.}
Deterministic checks address geometric failure modes; a separate,
checklist-based model review can assess visual discrepancies. The
maximum turn count and tool-call budget constrain the authoring session,
while a reserved final portion of calls directs completion of edits.
The reserve is configurable. Enforcing it by disabling exploratory tools
requires a pre-tool hook; adapters without this hook implement a hard
call-budget stop. In either case, the saved program must still be built
and checked, and termination is not a QC verdict.

\paragraph{Exporting the physical scene.}
The exporter combines four sources: shell geometry from \texttt{Room.py},
object poses and support declarations from \texttt{room\_layout.json},
appearance from \texttt{Room.glb}, and per-object physics records. Walls
are decomposed into structural boxes around door and window openings.
Attached fixtures are fixed, movable objects have free joints, and
articulated parts use their specified hinge or slide joints.

The transform between an asset's local geometry and its placed instance
also places its collision geometry and joint frames. When size changes,
derived mass scales with volume, authored mass is retained, and inertia
is recomputed from the placed geometry. Collider bounds are checked
against the recorded object bounds. Joints with compiled effort metadata
receive motor actuators; velocity limits are retained as controller
metadata. Newly authored objects without physics records use export-time
mass and collider estimates, and these fallbacks appear in the export
report.

Figure~\ref{fig:simulation} in the main paper illustrates the exported
scene under perturbation and robot interaction.

\clearpage
\subsection{Additional Scene-Editing Demonstrations}
\label{app:editing_demonstrations}
\begin{figure}[h]
  \centering
  \includegraphics[width=.95\linewidth]{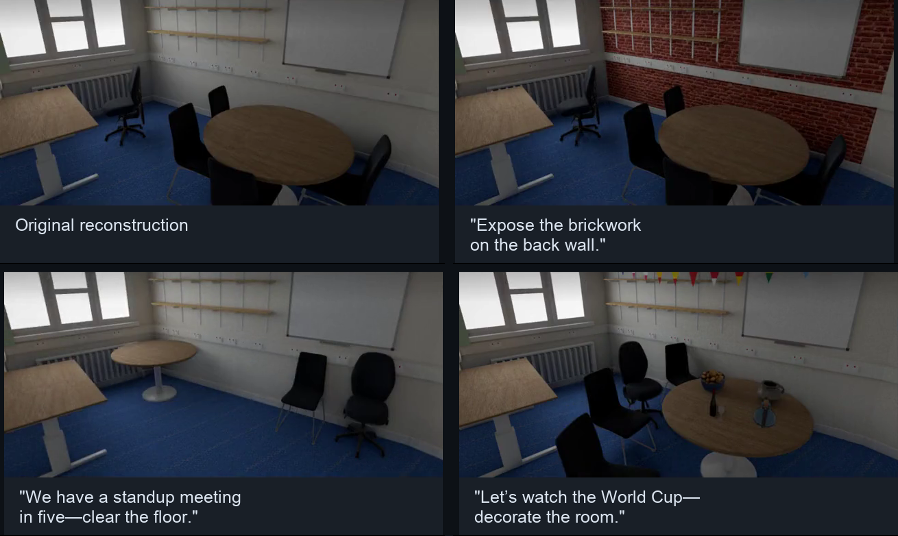}
  \caption{\textbf{Natural-language scene editing.} The original
  reconstruction is shown with three requested edits: expose the
  brickwork, clear the floor for a standup meeting, and decorate the
  room for watching the World Cup. Each instruction is applied as an
  edit to \texttt{Room.py}.}
  \label{fig:app_editing}
\end{figure}

Figure~\ref{fig:app_editing} shows natural-language editing examples
that complement the controlled programmatic edits in
Fig.~\ref{fig:interaction}. The requests change the wall material,
clear space for a meeting, and add decorations to the reconstructed office.

\paragraph{Controlled programmatic demonstrations.}
The variants in Fig.~\ref{fig:interaction} each start independently from
one saved 3D office scene and use an identical camera, resolution, and
render configuration. The following details describe these main-paper
examples.

\paragraph{Material editing.}
The carpet edit changes its colour while retaining the spatial texture,
room geometry, and furniture placement. This isolates an appearance
change from structural reconstruction.

\paragraph{Object removal.}
The layout edit removes the meeting-table group and four guest-chair
groups. Walls, shelves, and other fixtures remain in place, exposing the
floor previously occupied by the furniture.

\paragraph{Fixture addition.}
The lighting edit adds three pendant assemblies above the existing
table. Each assembly contains a shade, suspension cord, and light source,
with a ceiling-attachment declaration. The change affects both geometry
and illumination.

\paragraph{Provenance and scope.}
The images in Fig.~\ref{fig:interaction} are Blender renders of the
saved office reconstruction,
produced by a reproducible scene-editing script. They are qualitative
examples of what the editable representation permits, not additional
measurements of autonomous agent success or physical validity. The render
manifest records the source scene, camera, resolution, and sampling
settings for each variant.

%% file: sections/appendix/limitations.tex
\section{Limitations and Future Work}
\label{app:limitations}

\paragraph{Physical properties lack ground-truth validation.}
The physical properties used for simulation rely largely on
language-model inference and category or material priors, with
geometry-based estimates for quantities such as inertia. We do not have
ground-truth measurements of these properties for the reconstructed
objects. Simulation export and consistency checks therefore do not
establish that the scenes reproduce real-world dynamics. Collecting
measured physical properties and validating simulated behaviour against
real interactions are important future work.

\paragraph{Layout consistency does not guarantee accuracy.}
RoomPlan detections can contain errors in object dimensions and poses
that persist in the reconstructed layout. Our collision and support
corrections improve geometric consistency, but neither these checks nor
visual agreement with captured images guarantee recovery of the true
layout. Future work should more tightly integrate image evidence with
collision and support constraints during layout refinement, and evaluate
the resulting layouts against measured ground truth.

Broader evaluation should also cover larger and more structurally diverse
environments.